\pdfoutput=1
\documentclass[11pt,preprint]{article}

\usepackage[preprint]{acl}

\usepackage{times}
\usepackage{latexsym}

\usepackage[T1]{fontenc}
\usepackage[utf8]{inputenc}

\usepackage{microtype}

\usepackage{inconsolata}

\usepackage{graphicx}
\usepackage{xspace}

\usepackage{amsmath}
\usepackage{enumitem}
\usepackage{booktabs}
\usepackage{multirow}
\usepackage{url}
\usepackage{algorithmic}
\usepackage{algorithm}
\usepackage{multirow}
\usepackage{adjustbox}
\usepackage{wrapfig}

\newcommand{\oursystemname}{\textit{MCPEval}\xspace}

\title{\textit{MCPEval}: Automatic MCP-based Deep Evaluation for AI Agent Models}

\author{
  \textbf{Zhiwei Liu},
  \textbf{Jielin Qiu}, 
  \textbf{Shiyu Wang}, 
  \textbf{Jianguo Zhang}, 
  \textbf{Zuxin Liu}, 
  \textbf{Roshan Ram}, 
  \textbf{Haolin Chen}, 
    \\
  \textbf{Weiran Yao},
  \textbf{Shelby Heinecke}, 
  \textbf{Silvio Savarese},
  \textbf{Huan Wang}* \and
  \textbf{Caiming Xiong}* \\
  \\
  Salesforce AI Research \\
  *Corresponding Authors: \{huan.wang, cxiong\}@salesforce.com
}

\begin{document}
\maketitle
\begin{abstract}
The rapid rise of Large Language Models (LLMs)-based intelligent agents underscores the need for robust, scalable evaluation frameworks. Existing methods rely on static benchmarks and labor-intensive data collection, limiting practical assessment. We introduce \oursystemname, an open-source Model Context Protocol (MCP)-based framework that automates end-to-end task generation and deep evaluation of LLM agents across diverse domains. \oursystemname standardizes metrics, seamlessly integrates with native agent tools, and eliminates manual effort in building evaluation pipelines. Empirical results across five real-world domains show its effectiveness in revealing nuanced, domain-specific performance. 
We publicly release \oursystemname \footnote{\url{https://github.com/SalesforceAIResearch/MCPEval}} to promote reproducible and standardized LLM agent evaluation.

\end{abstract}

\section{Introduction}
\label{sec:introduction}

The rapid proliferation of Large Language Models (LLMs) has catalyzed a transformative shift in artificial intelligence, giving rise to autonomous agents capable of sophisticated reasoning, planning, and executing complex tasks \cite{openai2023gpt4,liu2024agentlite,liu2024practoptimizingprincipledreasoning,zhang2024xlam,ma2024taco}.
These systems, defined by their ability to autonomously conceive and adapt plans while interacting with dynamic environments through external tools, represent a new paradigm in AI \cite{ouyang2022training, zhang2023surveyagents}.
Consequently, there is an urgent and growing focus on developing comprehensive and systematic evaluation frameworks to standardize the assessment~\cite{zhu2024survey, zheng2024helm, ji2024safetybench}, thereby enabling the quick deployment and adaptation of these powerful systems. 

Current evaluation methods face significant limitations.
Early benchmarks, while foundational, often relied on static, predefined tasks, thus failing to capture the interactive real-world agentic workflows \cite{liu2023agentbench, liu2023bolaa, fan2024static}.
While the field has progressed toward dynamic and interactive benchmarks \cite{qin2023toolbench, zheng2023mtbench}, a persistent challenge remains: the lack of deep integration with practical tools and the absence of standardized protocols for agent-environment communication, which impedes reproducibility and robust comparison \cite{qin2023toolbench, deng2023mind2web, huang2024survey}.

\begin{figure*}[!ht]
  \centering
  \includegraphics[width=0.9\linewidth]{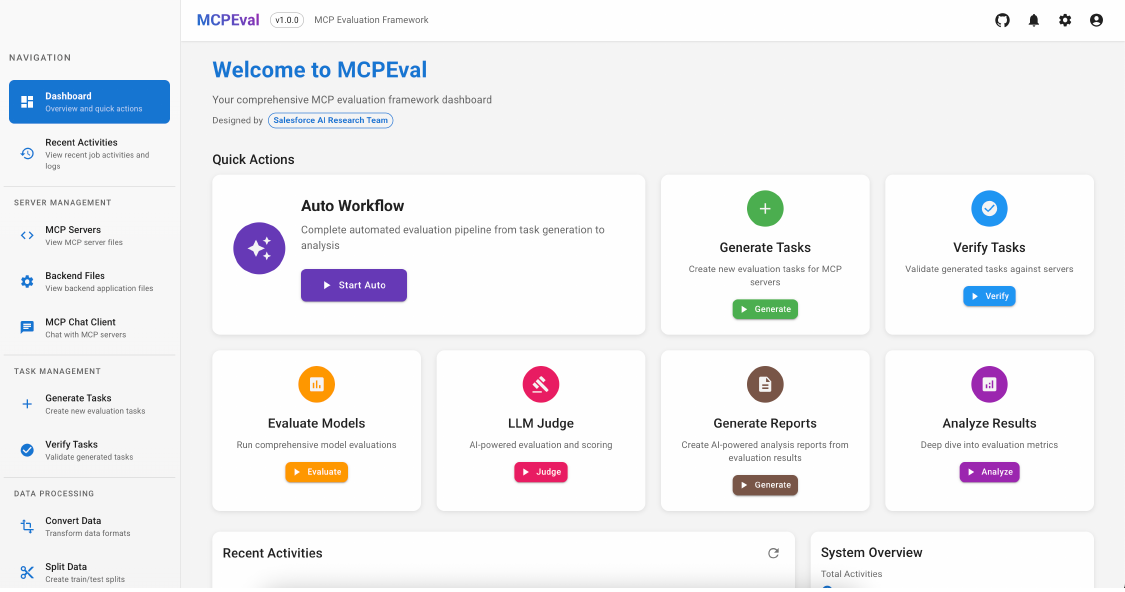}
  \caption{
    User interface of the MCPEval framework. The dashboard provides streamlined access to core functionalities such as automatic task generation, verification, model evaluation, and result analysis. It integrates real-time activity tracking and system overviews to ensure transparency and ease of use.
  }
  \label{fig:system_ui}
  \vspace{-5pt}
\end{figure*}

Addressing these critical gaps, we introduce \oursystemname, an automatic deep evaluation system for AI agents built upon the Model Context Protocol (MCP)\footnote{\url{https://modelcontextprotocol.io/introduction}}.
The recent emergence of MCP as a standard for governing interactions between LLMs and external systems offers a foundational layer for building scalable and interactive agents \cite{anthropic2024mcp, allganize2024mcp, lumer2025scalemcp}.
This standardization opens the quick development of new evaluation frameworks like MCP-Radar \cite{gao2025mcpradar} and MCPWorld \cite{yan2025mcpworld}.

Nevertheless, these approaches face significant scalability limitations, either requiring manual task creation and relies on labor intensive human evaluations for complex scenarios, or evaluated on static benchmark completion without comprehensive analysis.
\oursystemname advances beyond these systems by introducing a fully automated evaluation process. 
\oursystemname not only overcomes the manual bottlenecks and scalability issues in collecting tasks, but also provides a deep analysis workflow to evaluate models. 
As a result, our system can rapidly evaluate model's interactive capabilities towards new MCP tools and servers at scale, providing developers with actionable feedback to optimize their implementations.
The user interface design of \oursystemname is illustrated in Figure~\ref{fig:system_ui}.

\oursystemname goes beyond traditional success/failure metrics by systematically collecting detailed task trajectories and protocol interaction data, creating unprecedented visibility into agent behavior and generating valuable datasets for iterative improvement.
Additionally, because both task creation and verification are fully automated, the resulting high-quality trajectories can be immediately leveraged for rapid fine-tuning and continual improvement of agent models.
The comprehensive evaluation reports generated by \oursystemname also provide actionable insights towards the correctness of agent-platform communication at a granular level.

Through extensive experiments involving state-of-the-art LLMs and five real-world benchmarks, \oursystemname reveals scenarios where smaller, tool-enhanced models perform comparably to larger, more resource-intensive models.
These findings highlight significant opportunities for cost-effective deployments without compromising performance.
\oursystemname is made available as an open-source platform to support reproducibility, foster robust evaluation practices, and accelerate practical advancements in the broader LLM research community.

\vspace{-5pt}
\section{Related Work}
\label{sec:related}

This section reviews the shift in LLM evaluation from static benchmarks to dynamic, agent-based methods, highlighting frameworks like the Model Context Protocol (MCP) and the role of synthetic data in creating new test scenarios. 
% It identifies key gaps in current approaches, motivating the need for our framework.
%For comprehensive surveys on LLM agent evaluation, we refer readers to recent works by Huang et al.~\cite{huang2024survey} and Fan et al.~\cite{fan2024static}.
More comprehensive related work discussion can be found in Appendix~\ref{appendix:related-work}.

\vspace{-5pt}
\paragraph{The Evolution of LLM \& Agent Evaluation Frameworks}

LLM evaluation has evolved from static benchmarks like HELM~\cite{liang2022helm}, BIG-bench~\cite{srivastava2022beyond}, and MMLU~\cite{hendrycks2020measuring} to more interactive methods. To address data contamination, newer benchmarks employ strategies like temporal cutoffs, e.g., LiveBench~\cite{white2024livebench}) and LLM-based generation, e.g. MMLU-Pro~\cite{zhang2024mmlupro}.
Recognizing the limitations of single-turn formats, conversational and agentic benchmarks emerged, such as MT-Bench~\cite{zheng2023mtbench}, AgentBoard~\cite{ma2024agentboard}, and AgentBench~\cite{liu2023agentbench}, which focus on multi-turn agent performance. 
However, many still lack deep tool integration, prompting the development of domain-specific, task-oriented evaluations, such as  WebArena~\cite{xie2024osworld} and others~\cite{zhou2023webarena,liu2023bolaa,kokane2024spectool,koh2024visualwebarena,jimenez2023swe,geng2025realm,zhang2025actionstudio}.
\begin{figure*}[!htp]
\centering
\includegraphics[width=0.8\linewidth]{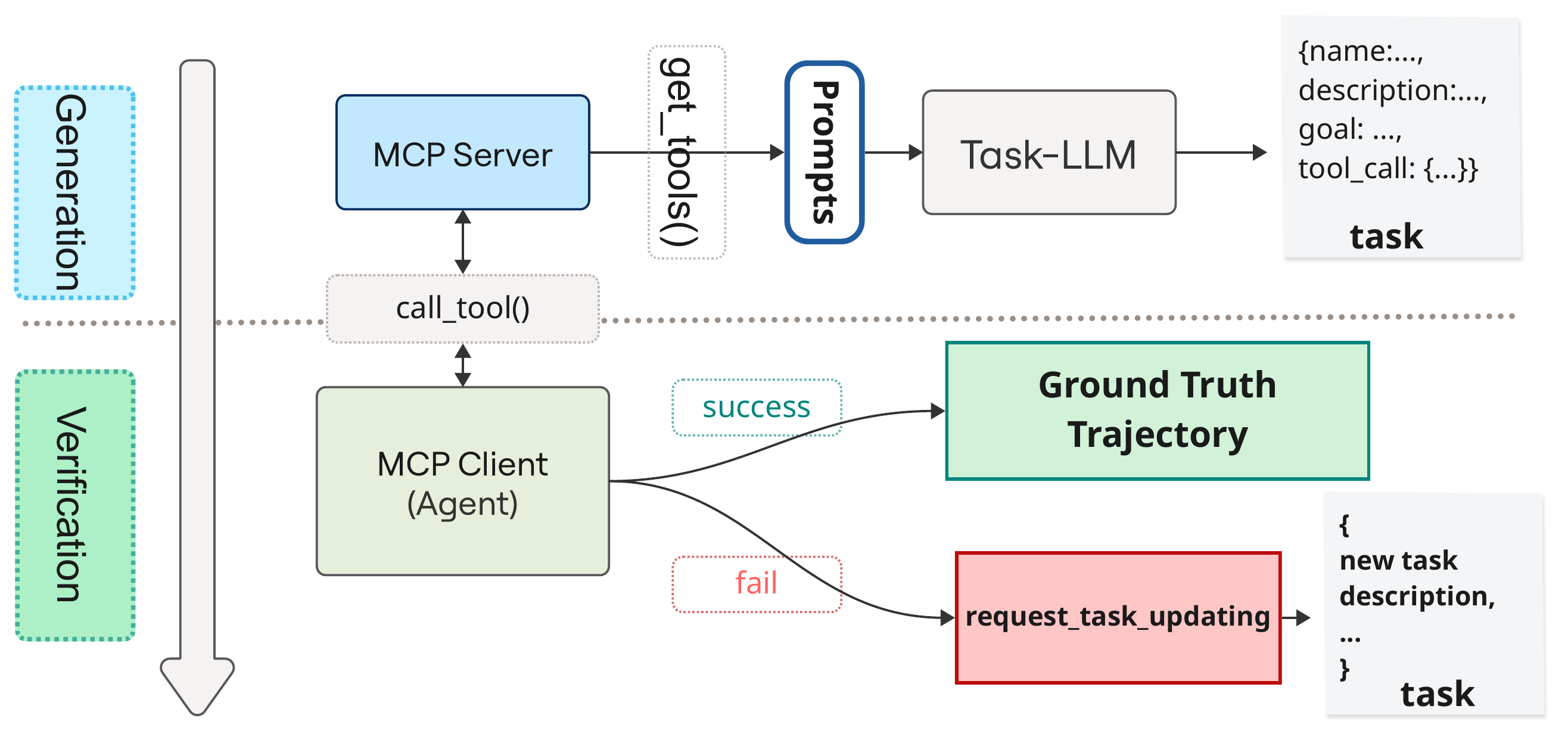}
\caption{Two-step MCP-based task generation workflow, including initial generation phase and verification phase. }
\label{fig:mcp_task_generation}
\vspace{-10pt}
\end{figure*}
\vspace{-5pt}
\paragraph{Evaluating Agents with Deep System Integration}

As agents increasingly operate in realistic digital environments, evaluation has shifted toward measuring system-level interaction. Platforms like OSWorld~\cite{xie2024osworld} assess GUI-based task execution, while frameworks like LangChain~\cite{langchain}, AutoGen~\cite{autogen}, and CrewAI~\cite{crewai} highlight the lack of robust evaluation tools. The MCP \cite{lumer2025scalemcp, allganize2024mcp} has emerged as a key standard for LLM-tool communication, enabling scalable, protocol-based evaluation. Frameworks like MCP-Radar~\cite{gao2025mcpradar} and MCPWorld~\cite{yan2025mcpworld} assess tool-use effectiveness and task completion. Our work builds on this by offering a finer-grained analysis of agent-protocol interaction fidelity.

\vspace{-5pt}
\paragraph{Synthetic Data Generation}

Recent work leverages LLMs to generate evaluation data, evolving from simple instruction-response pairs~\cite{zhang2023dialogstudio,wang2023selfinstruct,xu2023wizardlm,tan2025personabenchevaluatingaimodels} to rich and interactive scenarios~\cite{ma2024taco,tang2024matrix,liu2024apigen}.
Execution feedback enables closed-loop systems that automatically verify task correctness, as seen in AgentEval~\cite{arabzadeh2024agenteval} and LAMSimulator~\citep{hoang2025lam}. Our approach also uses synthetic data to evaluate agent interactions under MCP, generating tool-use tasks.
\section{\oursystemname Framework}
\label{sec:system_framework}
\oursystemname adopts a evaluation workflow designed with task generation, task verification and model evaluation, facilitating efficient and extensible evaluation of LLM agents. 
The task generation begins at the calling MCP server with  tool call method to collect the specifications of tools into prompts. We demonstrate this process in Figure~\ref{fig:mcp_task_generation}.

Upon receiving those context, a Task-LLM generates detailed task instructions, ensuring the appropriate information for tool calls are included in the task instruction.
However, we observe that these initial generated tasks may not include all necessary information to fill the tool parameters. 
Therefore, a task verification is necessary to generate high-quality task and collect the ground truth trajectory, which is illustrated in Figure~\ref{fig:mcp_task_generation}.

During verification stage, we employ a frontier agent as MCP client and interact the MCP server. 
This frontier agent executes the generated tasks with actual tool calls. Successful execution trajectory leads directly to verified tasks and corresponding ground truth. 
% Successful trajectories proceed directly to verified tasks and ground truth. 
In cases of task execution failure, the agent initiates a task updating request, prompting the generation of refined task descriptions. This iterative verification and refinement process ensures high-quality tasks suitable for comprehensive evaluation.
\begin{figure*}[]
\centering
\includegraphics[width=0.99\linewidth]{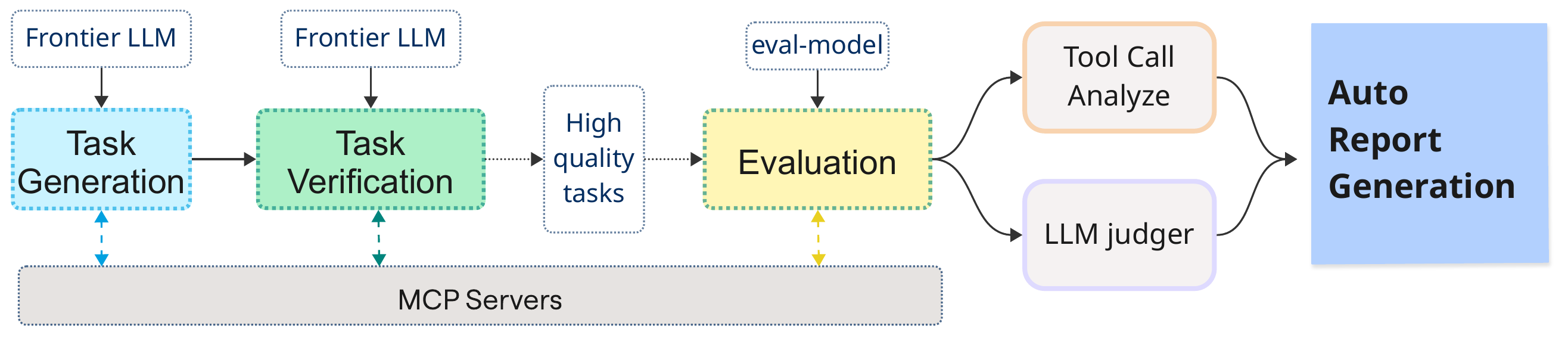}
\caption{\oursystemname evaluation workflow shows MCP client/server interaction, tool call correctness checking, LLM judger assessment, automated report generation.}
\label{fig:evaluation_framework}
\vspace{-10pt}
\end{figure*}

For model evaluation in Figure \ref{fig:evaluation_framework}, \oursystemname systematically assesses LLM agent models by placing the model-under-test as the MCP client and requiring it to complete the set of verified tasks. 
The framework then analyzes collected trajectories using two complementary perspectives: 1) Tool Call Matching, which rigorously compares the model’s tool usage against ground truth reference trajectories, and 
2) LLM Judging, which covers dimensions such as planning, execution flow, context awareness and etc.

Combining these analytic results, \oursystemname automatically generates comprehensive reports detailing each agent model’s strengths, weaknesses, and performance across multiple domains.
This fully automated workflow eliminates manual data collection and evaluation bottlenecks, enabling large-scale, reproducible agent assessment.

\section{Experiment}
\label{sec:experiments}

\vspace{-5pt}
\paragraph{Model Selection}
Our evaluation includes ten models spanning different architectures and capabilities, including seven \texttt{OpenAI Models}: GPT-4o, GPT-4o-mini, GPT-4.1-mini, GPT-4.1-nano, O3, O3-mini, O4-mini, and three \texttt{Open-Source Models}: Mistral-Small-24B, Qwen3-32B, Qwen3-30B-A3B. Model versions are listed in Table \ref{appendix-model-versions}.

\vspace{-5pt}
\paragraph{Domain Coverage}
We evaluate performance across five application domains: 
(1) \texttt{Healthcare}: Medical terminology lookup, drug information, clinical trials, health topics, PubMed search;
(2) \texttt{Airbnb}: Property search, listing details, booking information
(3) \texttt{Sports}: Team statistics, player information, game schedules, league standings;
(4) \texttt{National Parks}: Park information, visitor services, trail details, facility booking;
(5) \texttt{YFinance}: Stock prices, financial data, market analysis, portfolio management.

\begin{table*}[!ht]\small
\centering
\caption{Tool-call evaluation results: Strict and Flex tool matching scores across all MCP domains.}
\vspace{-5pt}
\label{tab:main_results}
\scriptsize
\begin{adjustbox}{width=0.95\linewidth}
\begin{tabular}{lcccccccccccccccc}
\toprule
\textbf{\multirow{2}{*}{Model}} &\multicolumn{2}{c}{\textbf{Finance}} &\multicolumn{2}{c}{\textbf{Airbnb}} &\multicolumn{2}{c}{\textbf{Healthcare}} &\multicolumn{2}{c}{\textbf{Sports}} &\multicolumn{2}{c}{\textbf{National Parks}} &\multicolumn{2}{c}{\textbf{Average}} \\
&\textbf{Strict} &\textbf{Flex} &\textbf{Strict} &\textbf{Flex} &\textbf{Strict} &\textbf{Flex} &\textbf{Strict} &\textbf{Flex} &\textbf{Strict} &\textbf{Flex} &\textbf{Strict} &\textbf{Flex} \\
\midrule
\textbf{mistral-small-24b} &42.0\% &43.0\% &65.3\% &72.3\% &82.0\% &87.0\% &64.0\% &66.0\% &52.0\% &62.0\% &61.1\% &66.1\% \\
\textbf{qwen3-30b-a3b} &38.0\% &38.0\% &63.0\% &69.0\% &80.0\% &85.0\% &72.0\% &74.0\% &50.0\% &57.0\% &60.6\% &64.6\% \\
\textbf{qwen3-32b} &51.0\% &52.0\% &63.0\% &70.0\% &82.0\% &87.0\% &70.0\% &72.0\% &54.0\% &63.0\% &64.0\% &68.8\% \\
\textbf{gpt4.1-mini} &92.0\% &93.0\% &75.0\% &\textbf{82.0\%} &\textbf{89.0\%} &\textbf{93.0\%} &82.0\% &84.0\% &59.0\% &66.0\% &79.4\% &83.6\% \\
\textbf{gpt4.1-nano} &86.0\% &88.0\% &64.4\% &68.7\% &70.0\% &75.0\% &73.3\% &75.0\% &50.0\% &59.0\% &68.7\% &73.1\% \\
\textbf{gpt4o-mini} &92.0\% &\textbf{94.0\%} &71.5\% &77.0\% &86.7\% &90.6\% &77.0\% &79.0\% &58.0\% &68.0\% &77.0\% &81.7\% \\
\textbf{gpt4o} &\textbf{93.0\%} &93.5\% &\textbf{77.0\%} &81.0\% &87.0\% &92.0\% &\textbf{80.0\%} &\textbf{82.0\%} &\textbf{64.0\%} &\textbf{73.0\%} &\textbf{80.2\%} &\textbf{84.3\%} \\
\textbf{o3-mini} &83.0\% &84.0\% &57.5\% &66.1\% &28.0\% &30.0\% &61.0\% &62.0\% &51.0\% &59.0\% &56.1\% &60.2\% \\
\textbf{o4-mini} &79.0\% &80.0\% &61.5\% &71.7\% &67.0\% &72.0\% &73.0\% &75.0\% &50.0\% &58.0\% &66.1\% &71.3\% \\
\textbf{o3} &86.0\% &87.0\% &66.0\% &79.0\% &62.0\% &69.0\% &68.0\% &70.0\% &44.0\% &54.0\% &65.2\% &71.8\% \\
\bottomrule
\end{tabular}
\end{adjustbox}
\vspace{-5pt}
\end{table*}
\begin{table*}[!]\small
\centering
\caption{LLM-Judge evaluation results: Trajectory (Traj) and Completion (Comp) across all MCP domains.}
\vspace{-5pt}
\label{tab:llm_judge_results}
\scriptsize
\begin{adjustbox}{width=0.95\linewidth}
\begin{tabular}{lccccccccccccccccc}\toprule
\textbf{\multirow{2}{*}{Model}} &\multicolumn{2}{c}{\textbf{Finance}} &\multicolumn{2}{c}{\textbf{Airbnb}} &\multicolumn{2}{c}{\textbf{Healthcare}} &\multicolumn{2}{c}{\textbf{Sports}} &\multicolumn{2}{c}{\textbf{National Parks}} &\multicolumn{2}{c}{\textbf{Average}}\\
&\textbf{Traj} &\textbf{Comp} &\textbf{Traj} &\textbf{Comp} &\textbf{Traj} &\textbf{Comp} &\textbf{Traj} &\textbf{Comp} &\textbf{Traj} &\textbf{Comp} &\textbf{Traj} &\textbf{Comp} \\
\midrule
\textbf{mistral-small-24b} &47.9\% &37.3\% &88.0\% &85.5\% &90.9\% &83.5\% &66.0\% &70.3\% &86.0\% &55.5\% &75.8\% &66.4\% \\
\textbf{qwen3-30b-a3b} &52.6\% &33.9\% &81.0\% &71.2\% &92.3\% &83.7\% &70.9\% &58.7\% &79.2\% &59.9\% &75.2\% &61.5\% \\
\textbf{qwen3-32b} &61.8\% &45.1\% &86.5\% &83.8\% &90.8\% &84.3\% &72.6\% &61.0\% &79.6\% &62.8\% &78.3\% &67.4\% \\
\textbf{gpt4.1-mini} &97.4\% &92.2\% &90.3\% &79.8\% &92.1\% &78.1\% &79.0\% &67.9\% &77.1\% &74.3\% &87.2\% &78.5\% \\
\textbf{gpt4.1-nano} &\textbf{97.7\%} &86.3\% &85.6\% &68.7\% &90.9\% &80.6\% &75.2\% &54.7\% &61.9\% &52.7\% &82.3\% &68.6\% \\
\textbf{gpt4o-mini} &96.7\% &92.4\% &90.2\% &82.4\% &\textbf{92.6\%} &78.2\% &80.1\% &68.7\% &83.0\% &77.0\% &88.5\% &79.7\% \\
\textbf{gpt4o} &97.0\% &92.3\% &91.1\% &75.3\% &92.4\% &79.9\% &80.8\% &68.7\% &\textbf{90.2\%} &87.1\% &\textbf{90.3\%} &80.7\% \\
\textbf{o3-mini} &63.6\% &60.6\% &91.4\% &92.4\% &77.9\% &54.1\% &86.8\% &80.0\% &90.1\% &\textbf{90.9\%} &82.0\% &75.6\% \\
\textbf{o4-mini} &37.2\% &36.8\% &91.6\% &93.6\% &85.8\% &92.3\% &\textbf{88.4\%} &88.2\% &81.3\% &86.6\% &76.9\% &79.5\% \\
\textbf{o3} &95.9\% &\textbf{97.1\%} &\textbf{92.3\%} &\textbf{97.4\%} &83.6\% &\textbf{94.8\%} &85.5\% &\textbf{94.5\%} &80.7\% &90.8\% &87.6\% &\textbf{94.9\%} \\
\bottomrule
\end{tabular}
\end{adjustbox}
\vspace{-5pt}
\end{table*}

\vspace{-5pt}
\paragraph{Evaluation Criteria}
Detailed explanation of those evaluation criteria are in Appendix~\ref{appendix:evaluation criteria}.
Our multi-level evaluation approach combines both tool call analyses and llm judger analyses over 676 tasks for all ten models:
(1) \texttt{Tool Call Analysis}: direct scoring of tool call matching, including parameter matching score, tool name matching score and tool order matching score. 
A overall matching score is calculated as a weighted combination of all.
We mainly report the overall scores in this paper.
We calculate both the strict and flexible matching scores.
(2) \texttt{LLM Judger Analysis}: Systematic evaluation of 50 model-domain combinations using frontier LLM as an expert judge. 
The judge rubrics are based on both trajectory-wise and final response completion-wise. 

\section{Results and Discussions}
\label{sec:results}

In this section, we introduce part of the experimental results and analysis from \oursystemname.
Appendix~\ref{appendix:results} includes more results and discussions.

\subsection{Cross-Domain Model Comparison}
We includes cross-domain model comparison results as in Table~\ref{tab:main_results} and Table~\ref{tab:llm_judge_results}, which respectively from tool-call criteria and LLM judge criteria. 

Regarding the tool-call performance, Table~\ref{tab:main_results} shows that GPT-4 variants, especially \texttt{gpt4o}, consistently lead across domains.
Open models display mixed performance.
Notably, smaller models like \texttt{o4-mini} generalize well, outperforming larger open models in certain tasks. 
These findings highlight the value of \oursystemname in revealing nuanced, domain-specific strengths and weaknesses that static benchmarks may overlook.

Beyond binary success rates, Table~\ref{tab:llm_judge_results} uses LLM-Judge to assess interaction quality through trajectory (reasoning) and completion (task alignment) scores. GPT-4 models again lead, with \texttt{o3} excelling in Completion and \texttt{gpt4o-mini} and \texttt{gpt4.1-mini} showing strong Trajectory performance. 
Open models perform inconsistently, where some succeed in domains like Healthcare but falter in reasoning stability. 
Notably, \texttt{o4-mini} often recovers from weak reasoning to deliver strong final outputs.
These results show how \oursystemname paired with LLM-based evaluation reveals deeper insights into agent behavior across tasks.

\subsection{Fine-grained Criteria Comparison}
\begin{table*}
    \centering
    \caption{LLM Judger and Tool Usage Evaluation Metrics (mean ± std).}
    \adjustbox{width=\textwidth,center}
    {
    \begin{tabular}{lccc|cccc}
    \toprule
    & \multicolumn{3}{c}{\textbf{Tool Call Metrics}} & \multicolumn{4}{c}{\textbf{LLM Judger Metrics}} \\
    \cmidrule(lr){2-4} \cmidrule(lr){5-8}
    \textbf{Model} & \textbf{Name Match} & \textbf{Param Match} & \textbf{Order Match} & \textbf{Planning} & \textbf{Execution} & \textbf{Req. Coverage} & \textbf{Completeness} \\
    \midrule
    \textbf{mistral-small-24b} & 0.548 ± 0.185 & 0.636 ± 0.165 & 0.645 ± 0.226 & 0.764 ± 0.176 & 0.723 ± 0.171 & 0.569 ± 0.148 & 0.509 ± 0.135 \\
    \textbf{qwen3-30b-a3b} & 0.564 ± 0.162 & 0.731 ± 0.200 & 0.650 ± 0.213 & 0.790 ± 0.133 & 0.726 ± 0.147 & 0.649 ± 0.188 & 0.577 ± 0.181 \\
    \textbf{qwen3-32b} & 0.592 ± 0.150 & 0.782 ± 0.117 & 0.693 ± 0.162 & 0.819 ± 0.104 & 0.757 ± 0.116 & 0.710 ± 0.176 & 0.638 ± 0.176 \\
    \textbf{gpt4.1-mini} & 0.768 ± 0.163 & \textbf{0.878 ± 0.087} & \textbf{0.887 ± 0.074} & 0.875 ± 0.083 & 0.849 ± 0.097 & 0.808 ± 0.100 & 0.742 ± 0.105 \\
    \textbf{gpt4.1-nano} & 0.630 ± 0.106 & 0.825 ± 0.130 & 0.738 ± 0.109 & 0.837 ± 0.131 & 0.801 ± 0.145 & 0.713 ± 0.168 & 0.633 ± 0.161 \\
    \textbf{gpt4o-mini} & 0.754 ± 0.157 & 0.862 ± 0.074 & 0.849 ± 0.100 & 0.895 ± 0.058 & 0.867 ± 0.074 & 0.832 ± 0.084 & 0.770 ± 0.095 \\
    \textbf{gpt4o} & \textbf{0.793 ± 0.106} & 0.876 ± 0.086 & 0.873 ± 0.081 & 0.902 ± 0.053 & \textbf{0.884 ± 0.064} & 0.838 ± 0.101 & 0.767 ± 0.119 \\
    \textbf{o3-mini} & 0.511 ± 0.196 & 0.688 ± 0.216 & 0.605 ± 0.191 & 0.856 ± 0.121 & 0.824 ± 0.111 & 0.795 ± 0.170 & 0.737 ± 0.177 \\
    \textbf{o4-mini} & 0.571 ± 0.113 & 0.818 ± 0.137 & 0.781 ± 0.082 & 0.783 ± 0.234 & 0.747 ± 0.216 & 0.819 ± 0.247 & 0.790 ± 0.236 \\
    \textbf{o3} & 0.588 ± 0.186 & 0.803 ± 0.116 & 0.803 ± 0.134 & \textbf{0.903 ± 0.042} & 0.858 ± 0.067 & \textbf{0.969 ± 0.029} & \textbf{0.955 ± 0.029} \\
    \bottomrule
    \end{tabular}\label{table:llm-judge-eval-metrics}
    }
    \end{table*}
Besides the cross-domain comparison, we also demonstrate the performance comparison of different models w.r.t. part of those fine-grained criteria aspects, including \textit{Name Match}, \textit{Parameter Match}, \textit{Order Match}, \textit{Planning Execution}, \textit{Requirements Coverage}, and \textit{Completeness} on both tool call metrics and LLM Judger Metrics.

Looking at the evaluation results from our \oursystemname framework, several key patterns emerge that must be interpreted within the context of the evaluation methodology. 
Since the tool call ground truth is generated from gpt-4.1 and other models are evaluated against this reference, the tool call metrics (Name Match, Param Match, Order Match) in Table \ref{table:llm-judge-eval-metrics} reflect alignment with gpt-4.1's specific approach rather than absolute tool calling quality.
Based on the observation of gpt4.1-mini achieving the highest scores in parameter matching (0.878) and order matching (0.887), we hypothesis that gpt-4.1-mini is a distilled model from gpt-4.1.

Conversely, the o3 model's lower performance in tool call precision metrics (Name Match: 0.588, Param Match: 0.803) may not indicate inferior tool calling capabilities, but rather a fundamentally different approach to API interaction that diverges from gpt-4.1's style.
The LLM judger metrics provide a more objective assessment of actual task performance, revealing o3's exceptional capabilities in planning (0.903), requirement coverage (0.969), and completeness (0.955). This suggests that while o3 may employ different tool calling patterns compared to gpt-4.1, it demonstrates superior high-level reasoning and task comprehension. The balanced performance of gpt4o and gpt4o-mini across both metric categories indicates these models successfully combine effective planning with tool calling approaches that align well with the gpt-4.1 reference standard. For open-source models like mistral-small-24b and the qwen3 variants, the lower scores across all metrics reflect both genuine performance gaps and potential stylistic differences in tool usage patterns compared to the proprietary reference model, highlighting the challenge of fair evaluation when using model-specific ground truth standards.

\subsection{Domain Performance Analysis}

\begin{figure}[tbp]
\centering
\includegraphics[width=0.99\linewidth]{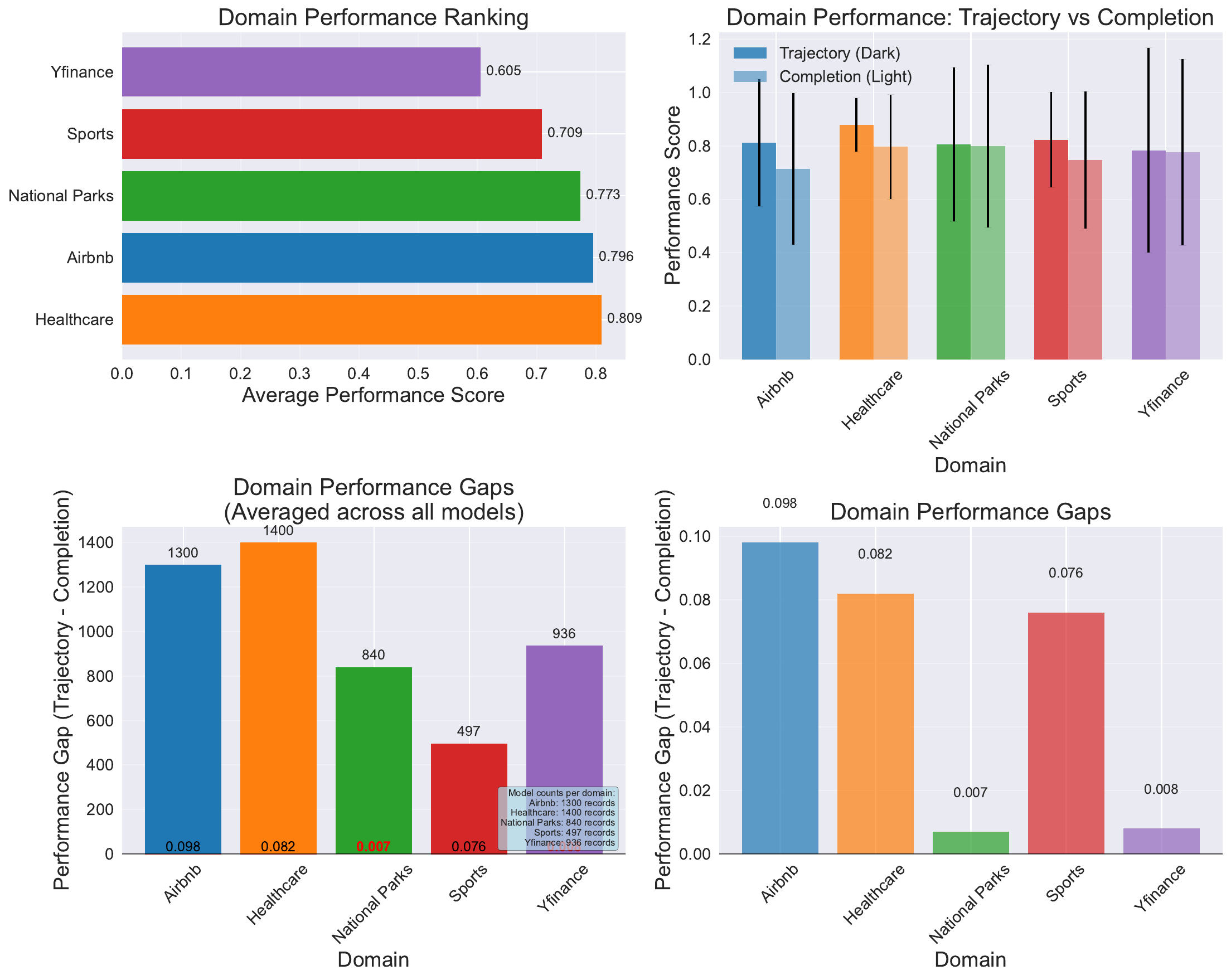}
\caption{Domain performance analysis: (a) Domain ranking by LLM judger, (b) Trajectory vs completion comparison, (c) Task distribution, (d) Performance gaps by domain.}
\label{fig:domain_analysis_main}
\vspace{-20pt}
\end{figure}

Figure~\ref{fig:domain_analysis_main} presents comprehensive domain performance analysis across multiple dimensions, revealing significant variations in task complexity and tool call quality across different domains.
Figure~\ref{fig:domain_analysis_main} reveals a domain performance hierarchy led by Healthcare, followed by Airbnb, Sports, Finance, and National Parks, driven by API structure and task complexity. Healthcare excels due to standardized data and strong alignment between API outputs and user needs, while National Parks struggles with complex geographical data integration. Execution vs. completion patterns vary, where Finance stands out for strong completions despite middling trajectories, and Airbnb shows a large gap, highlighting difficulty in generating comprehensive results. Task distribution is highest in Healthcare and Airbnb, ensuring robust analysis. Overall, domain-specific gaps reflect differences in API design quality and task demands.

\subsection{Performance Gap Analysis}

\begin{figure}[tbp]
\centering
\includegraphics[width=0.99\linewidth]{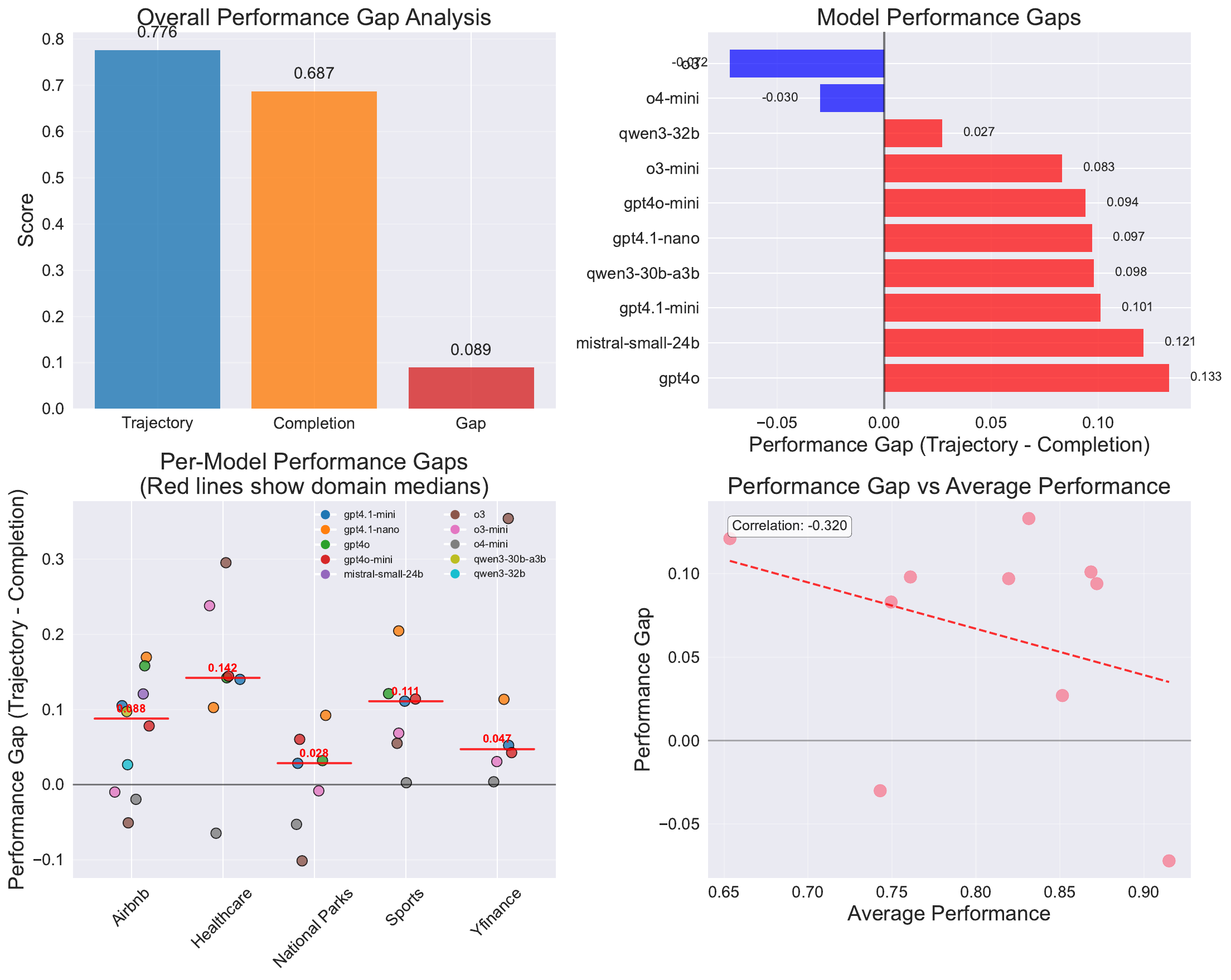}
\caption{Performance gap analysis: (a) Overall gap distribution, (b) Model-wise gaps, (c) Domain-wise gaps, (d) Gap-performance correlation.}
\label{fig:gap_analysis_main}
\vspace{-10pt}
\end{figure}

Figure~\ref{fig:gap_analysis_main} presents comprehensive analysis of performance gaps across models and domains, revealing fundamental patterns in LLM tool-use capabilities and architectural characteristics.
Figure~\ref{fig:gap_analysis_main} reveals a consistent trend: models generally perform better at trajectory execution than completion, highlighting a maturity gap in synthesis capabilities. Most models show positive execution-completion gaps, with open-source models displaying larger gaps than OpenAI models. Notably, O3 exhibits a rare negative gap, excelling in completion quality. Domain-level analysis shows similar trends, with Finance having the smallest gap and Airbnb the largest, reflecting differences in API structure and task complexity. A weak negative correlation between performance and gap size suggests that stronger models tend to be more balanced, though architecture-specific factors also play a role.

% \subsection{Tool Usage Pattern Analysis}

% Table~\ref{tab:tool_patterns_main} reveals key limitations in LLMs’ ability to interact with APIs. Models often generate approximate but imprecise parameters, exposing a core weakness in semantic mapping and parameter specification. Multi-tool coordination further challenges models due to increased complexity in planning and context management.

% \begin{table}[tbp]\small
% \centering
% \caption{Tool usage pattern analysis.}
% \vspace{-5pt}
% \label{tab:tool_patterns_main}
% \begin{adjustbox}{width=0.99\linewidth}
% \begin{tabular}{lcc}
% \toprule
% \textbf{Pattern} & \textbf{Range} & \textbf{Observation} \\
% \midrule
% Exact Match & 0.027 - 0.695 & Wide performance variation \\
% Flexible Match  & 0.015 - 0.695 & Improved with flexibility \\
% Parameter Mismatches & Universal & Common across all models \\
% Multi-tool Coordination & Lower success & Complex task challenges \\
% Tool Combination Errors & Frequent & Sequence and dependency issues \\
% \bottomrule
% \end{tabular}
% \end{adjustbox}
% \vspace{-10pt}
% \end{table}

\subsection{Model Performance Hierarchy}
\begin{figure*}[tbp]
\centering
\includegraphics[width=0.99\linewidth]{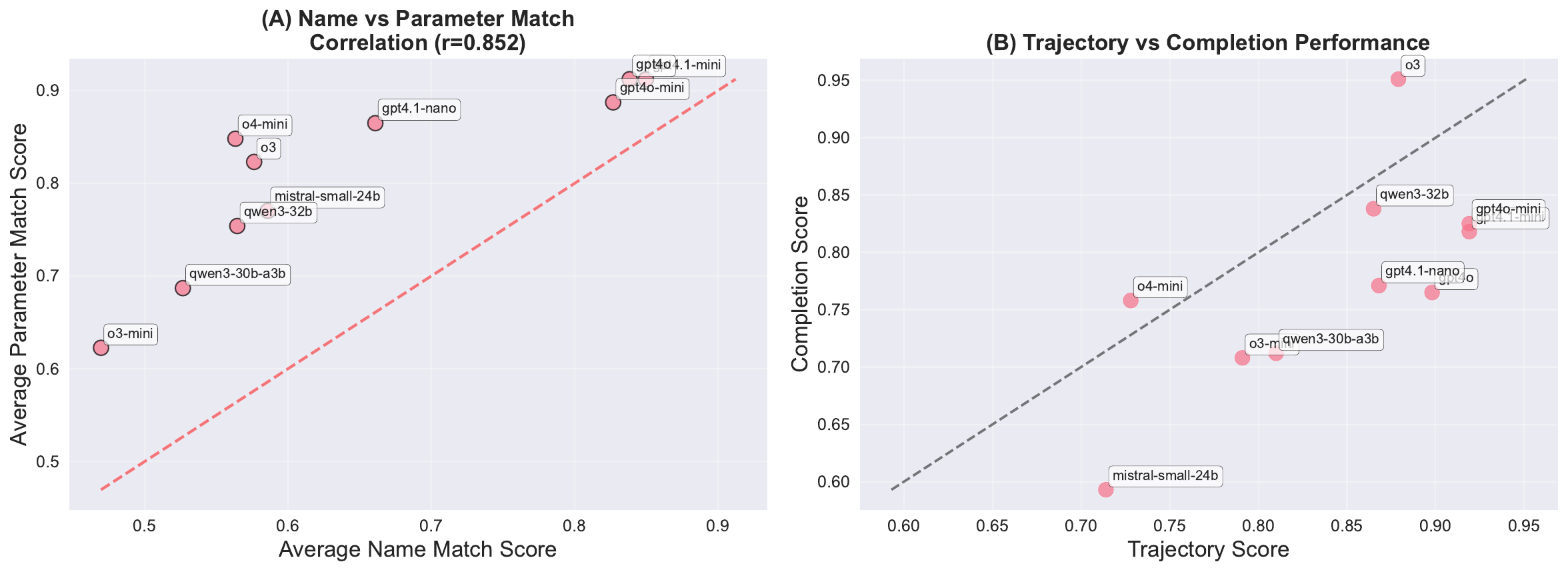}
\caption{Model performance analysis from tool call anlysis and llm judger.}
\label{fig:tool_analysis_comparison}
\vspace{-10pt}
\end{figure*}

Figure~\ref{fig:tool_analysis_comparison} illustrates the model performance 2-d comparison on both tool call and LLM judger. 
Regarding tool call criteria, we use name matching and parameter matching scores, while LLM judger criteria averages the scores w.r.t. trajectory and completion.
It highlights a clear performance hierarchy among models, with OpenAI's O3, GPT-4o-mini, and GPT-4.1-mini consistently leading across metrics. OpenAI models outperform open-source ones in overall quality with lower variance, indicating more stable tool-use capabilities.  
Additionally, the higher parameter match scores than the name match score indicate that models understand the context but may select different tool compared with ground truth from GPT-4.1.  
A trade-off between trajectory execution and completion quality is evident, with most models excelling in execution but not completion, except O3, which uniquely excels in completion. The top models also show specialization, underscoring that strength in one area does not imply dominance in another, reflecting varied optimization strategies.

\subsection{Performance Correlation Analysis}

%\begin{figure}[htbp]
%\centering
%\includegraphics[width=0.5\linewidth]{figs/corr.pdf}
%\caption{Trajectory vs Completion Performance Correlation Across Domains.}
%\label{fig:corr_main}
%\vspace{-5pt}
%\end{figure}

% \begin{wrapfigure}[13]{r}{.3\textwidth}
%     %\vspace{-.5cm}
%     \centering
%     \includegraphics[width=0.99\linewidth]{figs/corr_v2.pdf}
%     %\vspace{-0.5cm}
%     \captionof{figure}{Trajectory vs Completion performance correlation across domains.}
%     \label{fig:corr_main}
%     %\vspace{-0.5cm}
% \end{wrapfigure}

\begin{figure}
    \centering
    \includegraphics[width=0.8\linewidth]{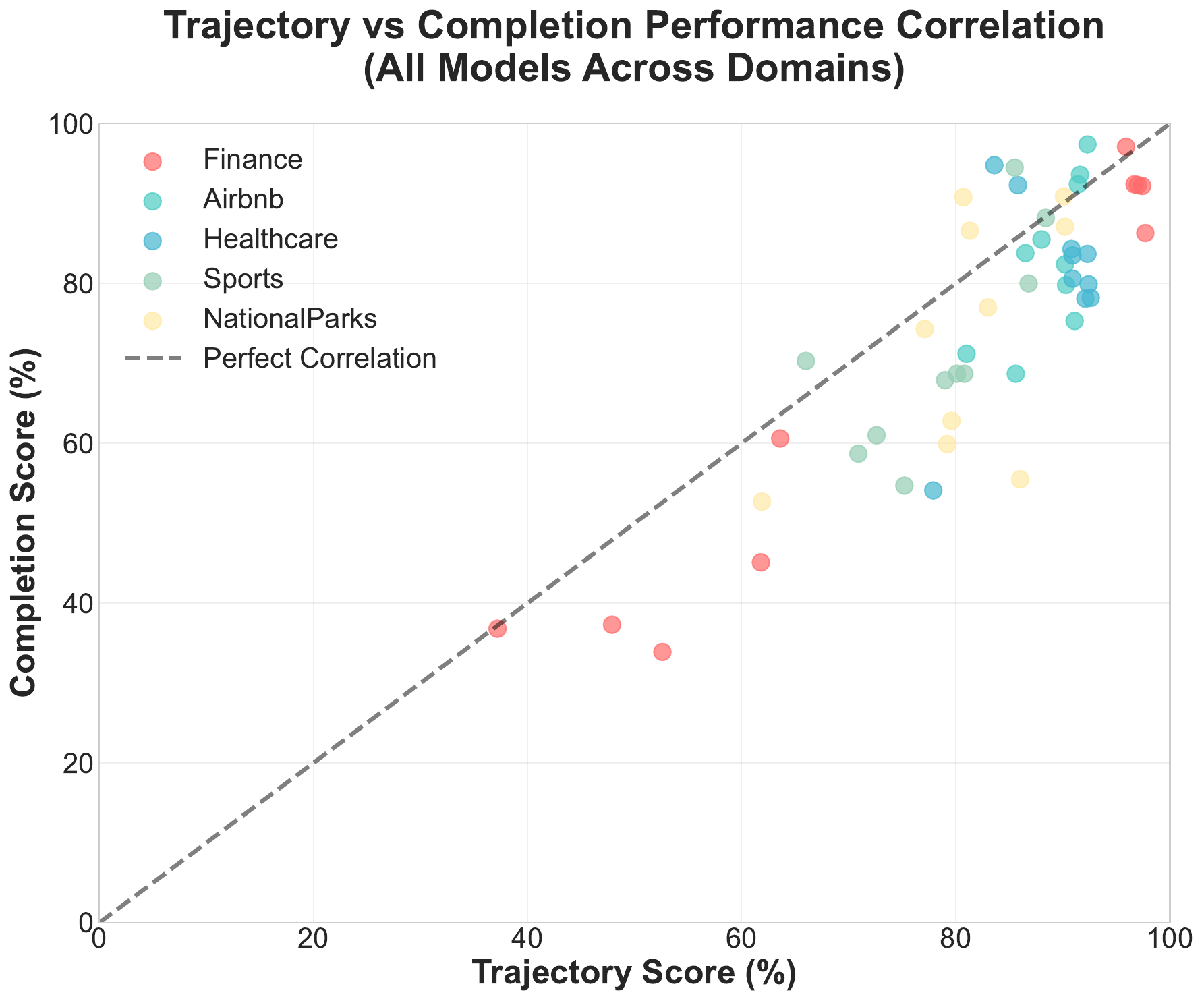}
    %\vspace{-0.5cm}
    \captionof{figure}{Trajectory vs Completion performance correlation across domains.}
    \label{fig:corr_main}
    %\vspace{-0.5cm}
\end{figure}

Figure~\ref{fig:corr_main} shows the universal execution-completion gap across all LLM models and domains. Finance shows the most pronounced gap, with models achieving 95\%+ trajectory scores but dropping to 85-95\% completion quality. Healthcare exhibits the most balanced performance near the diagonal line, while Airbnb shows high variability and Sports clusters in the middle range. National Parks demonstrates compressed performance in the lower range, reflecting complex geographical data integration challenges. Notably, no points appear above the diagonal reference line, confirming that current LLM consistently excel at procedural reasoning and tool execution but struggle with output synthesis across all domains.

\section{Conclusion}
\label{sec:conclusion}

We present \oursystemname, an automated, MCP-based evaluation framework designed to address critical gaps in LLM agent assessment. Our approach structured the evaluation process across diverse domains, automates end-to-end task generation, verification, and deep assessment,
eliminating the boundary in data collection and manual pipeline construction.
Extensive experiments across five real-world domains demonstrate that \oursystemname uncovers nuanced, domain-specific performance differences and provides actionable insights beyond traditional evaluation metrics. By releasing \oursystemname as an open-source toolkit, we aim to promote reproducible, scalable, and standardized evaluation practices for the LLM agent community and develop more robust and capable AI agents.

% \clearpage

\section*{Broader Impact}

\oursystemname has the potential for significant broader impact across multiple stakeholder communities. The framework serves the research community by providing a standardized platform for reproducible agent evaluation research, enabling consistent comparison and validation of different approaches. For industry adoption, it enables systematic assessment of agent readiness for production deployment, helping organizations make informed decisions about agent deployment strategies. In terms of model development, the framework informs training and development priorities through comprehensive capability analysis, guiding future research directions and resource allocation. Finally, the framework contributes to user safety by enabling thorough pre-deployment evaluation, helping ensure that deployed agents meet safety and reliability standards.

\section*{Limitations}
\label{sec:limitations}

Despite the strengths of \oursystemname, several limitations remain. First, our evaluation relies entirely on synthetic data, which may not fully reflect the complexity and unpredictability of real-world agent interactions.
Second, using LLM-based judges for long trajectories can be costly in terms of both computation and resource usage, potentially limiting scalability for large-scale or lengthy evaluations. 
Third, the automated verification process can introduce bias or produce false ground truth labels, especially for ambiguous or open-ended tasks, which may affect the reliability of certain evaluation results.
Future work includes incorporating real-world task data, developing more efficient and cost-effective judging methods, and improving verification strategies to reduce bias and enhance the accuracy of ground truth labels.
It would be a future patch to conduct verification and judgment from multiple sources for cross validation.

\section*{Acknowledgments}

We thank the open-source community for providing useful tools and MCP servers for our \oursystemname.
We also acknowledge the Model Context Protocol community and the various benchmark datasets that have enabled this work. And we would acknowledge part of code of the system are generated from multiple frontier AI models and verified with human checking. 

\bibliography{reference}

\begin{thebibliography}{48}
\providecommand{\natexlab}[1]{#1}

\bibitem[{Allganize(2024)}]{allganize2024mcp}
Allganize. 2024.
\newblock Alli for enterprise: On-premise llm app server \& the model context protocol (mcp).
\newblock Accessed: 2025-06-27.

\bibitem[{Anthropic(2024)}]{anthropic2024mcp}
Anthropic. 2024.
\newblock Model context protocol.
\newblock \url{https://github.com/modelcontextprotocol/}.
\newblock Accessed: 2024-06-26.

\bibitem[{Arabzadeh et~al.(2024)Arabzadeh, A.~D. J.~C., F.~C., Ipeirotis, Z., P., and S.}]{arabzadeh2024agenteval}
Negin Arabzadeh, Anand A.~D. J.~C., Fabio F.~C., P.~G. Ipeirotis, Jin Z., Panos P., and Sanmi S. 2024.
\newblock {AgentEval} 1.0: A comprehensive benchmark for evaluating autonomous agents.
\newblock \emph{arXiv preprint arXiv:2401.07303}.

\bibitem[{Chase et~al.(2022)}]{langchain}
Harrison Chase and 1 others. 2022.
\newblock {LangChain}.
\newblock \url{https://github.com/langchain-ai/langchain}.

\bibitem[{Deng et~al.(2023)Deng, Viguier, Chen, Gu, Zhang, Yogatama, Dréze, Jia, and Wang}]{deng2023mind2web}
Xiang Deng, Adrien Viguier, Xinyi Chen, C~Gu, Xinyun Zhang, D~Yogatama, M~Dréze, C~Jia, and W~Wang. 2023.
\newblock Mind2web: Towards a generalist agent for the web.
\newblock In \emph{Advances in Neural Information Processing Systems}.

\bibitem[{Fan et~al.(2024)Fan, Lin, Zhang, Yao, Zhu, Li, Qian, Jiang, Chen, Cheng et~al.}]{fan2024static}
Yapei Fan, Tiezheng Lin, Yutao Zhang, Qing Yao, Bo~Zhu, Yilun Li, Wenlin Qian, Xin Jiang, Wei Chen, Peiyi Cheng, and 1 others. 2024.
\newblock From static to dynamic: A survey of evaluation methods for large language models.
\newblock \emph{arXiv preprint arXiv:2402.04337}.

\bibitem[{Gao et~al.(2025)Gao, Wang, Kim, Lee, and Yang}]{gao2025mcpradar}
Xuanqi Gao, Yue Wang, Jinsu Kim, Chang Lee, and Diyi Yang. 2025.
\newblock {MCP-RADAR}: A multi-dimensional benchmark for evaluating tool use capabilities in large language models.
\newblock \emph{arXiv preprint arXiv:2505.16700}.

\bibitem[{Geng and Chang(2025)}]{geng2025realm}
Zhaolin Geng and Kewei Chang. 2025.
\newblock {REALM-Bench}: A real-world planning benchmark for llms and multi-agent systems.
\newblock \emph{arXiv preprint arXiv:2502.18836}.

\bibitem[{Hendrycks et~al.(2020)Hendrycks, Burns, Basart, Zou, Mazeika, Song, and Steinhardt}]{hendrycks2020measuring}
Dan Hendrycks, Collin Burns, Steven Basart, Andy Zou, Mantas Mazeika, Dawn Song, and Jacob Steinhardt. 2020.
\newblock Measuring massive multitask language understanding.
\newblock \emph{arXiv preprint arXiv:2009.03300}.

\bibitem[{Hoang et~al.(2025)Hoang, Huang, Kokane, Zhang, Liu, Zhu, Grigsby, Lan, Ryoo, Wu et~al.}]{hoang2025lam}
Thai Hoang, Kung-Hsiang Huang, Shirley Kokane, Jianguo Zhang, Zuxin Liu, Ming Zhu, Jake Grigsby, Tian Lan, Michael~S Ryoo, Chien-Sheng Wu, and 1 others. 2025.
\newblock Lam simulator: Advancing data generation for large action model training via online exploration and trajectory feedback.
\newblock \emph{arXiv preprint arXiv:2506.02298}.

\bibitem[{Huang et~al.(2024)Huang, Zeng, Chen, Wei, Yuan, Li, Yang, Wang, Liu, Wei et~al.}]{huang2024survey}
Zeshan Huang, Zihan Zeng, Keren Chen, Yihui Wei, Guilei Yuan, Jinhao Li, Jiaan Yang, Ziqi Wang, Jiateng Liu, Zexin Wei, and 1 others. 2024.
\newblock A survey on evaluation of large language models as agents.
\newblock \emph{arXiv preprint arXiv:2406.03456}.

\bibitem[{Ji et~al.(2024)Ji, Liu, Zhao, Yuan, Li, Lin, Wang, Zhang, and Liu}]{ji2024safetybench}
Zhexin Ji, Zhaofan Liu, Zihan Zhao, Fuhao Yuan, Cheng Li, Yang Lin, Pinyu Wang, Yaodong Zhang, and Jing Liu. 2024.
\newblock \href {https://aclanthology.org/2024.naacl-long.412} {{S}afety{B}ench: A comprehensive benchmark to evaluate {LLM}s' safety}.
\newblock In \emph{Proceedings of the 2024 Conference of the North American Chapter of the Association for Computational Linguistics: Human Language Technologies}, pages 7406--7421, Mexico City, Mexico. Association for Computational Linguistics.

\bibitem[{Jimenez et~al.(2023)Jimenez, He, Joty, and Shang}]{jimenez2023swe}
Carlos~E Jimenez, John He, Shafiq Joty, and Wei-Hao Shang. 2023.
\newblock Swe-bench: Can language models solve real-world software engineering problems?
\newblock \emph{arXiv preprint arXiv:2310.06770}.

\bibitem[{Koh et~al.(2024)Koh, Gu, Lee, Zhou, Geng, Zhu, Li, Gu, Zhang, Zhang et~al.}]{koh2024visualwebarena}
Jing~Yu Koh, Robert Gu, Hong-Lek Lee, Xuhui Zhou, Xingyu Geng, Hao Zhu, Zhengyun Li, Peiran Gu, Yin-Dong Zhang, Yi~Zhang, and 1 others. 2024.
\newblock {VisualWebArena}: A realistic and challenging benchmark for multimodal web agents.
\newblock \emph{arXiv preprint arXiv:2401.13649}.

\bibitem[{Kokane et~al.(2024)Kokane, Zhu, Awalgaonkar, Zhang, Hoang, Prabhakar, Liu, Lan, Yang, Tan et~al.}]{kokane2024spectool}
Shirley Kokane, Ming Zhu, Tulika Awalgaonkar, Jianguo Zhang, Thai Hoang, Akshara Prabhakar, Zuxin Liu, Tian Lan, Liangwei Yang, Juntao Tan, and 1 others. 2024.
\newblock Spectool: A benchmark for characterizing errors in tool-use llms.
\newblock \emph{arXiv preprint arXiv:2411.13547}.

\bibitem[{Liang et~al.(2022)Liang, Bommasani, Lee, Mada, Hudson, Hall, Icard, Adel, Adipo, Aina et~al.}]{liang2022helm}
Percy Liang, Rishi Bommasani, Tony Lee, D~Mada, D~Hudson, E~Hall, T~Icard, H~Adel, A~Adipo, J~Aina, and 1 others. 2022.
\newblock Holistic evaluation of language models.
\newblock In \emph{Advances in Neural Information Processing Systems}.

\bibitem[{Liu et~al.(2023{\natexlab{a}})Liu, Yu, Zhang, Xu, Wang, Zhang, Tan, Xu, Li, Yang et~al.}]{liu2023agentbench}
Xiao Liu, Hao Yu, Hanchen Zhang, Yaran Xu, Zekun Wang, Ruobing Zhang, C~Tan, C~Xu, X~Li, R~Yang, and 1 others. 2023{\natexlab{a}}.
\newblock Agentbench: Evaluating llms as agents.
\newblock \emph{arXiv preprint arXiv:2308.03688}.

\bibitem[{Liu et~al.(2024{\natexlab{a}})Liu, Yao, Zhang, Murthy, Yang, Liu, Lan, Zhu, Tan, Kokane, Hoang, Niebles, Heinecke, Wang, Savarese, and Xiong}]{liu2024practoptimizingprincipledreasoning}
Zhiwei Liu, Weiran Yao, Jianguo Zhang, Rithesh Murthy, Liangwei Yang, Zuxin Liu, Tian Lan, Ming Zhu, Juntao Tan, Shirley Kokane, Thai Hoang, Juan~Carlos Niebles, Shelby Heinecke, Huan Wang, Silvio Savarese, and Caiming Xiong. 2024{\natexlab{a}}.
\newblock \href {https://arxiv.org/abs/2410.18528} {Pract: Optimizing principled reasoning and acting of llm agent}.
\newblock \emph{Preprint}, arXiv:2410.18528.

\bibitem[{Liu et~al.(2023{\natexlab{b}})Liu, Yao, Zhang, Xue, Heinecke, Murthy, Feng, Chen, Niebles, Arpit et~al.}]{liu2023bolaa}
Zhiwei Liu, Weiran Yao, Jianguo Zhang, Le~Xue, Shelby Heinecke, Rithesh Murthy, Yihao Feng, Zeyuan Chen, Juan~Carlos Niebles, Devansh Arpit, and 1 others. 2023{\natexlab{b}}.
\newblock Bolaa: Benchmarking and orchestrating llm-augmented autonomous agents.
\newblock \emph{arXiv preprint arXiv:2308.05960}.

\bibitem[{Liu et~al.(2024{\natexlab{b}})Liu, Yao, Zhang, Yang, Liu, Tan, Choubey, Lan, Wu, Wang et~al.}]{liu2024agentlite}
Zhiwei Liu, Weiran Yao, Jianguo Zhang, Liangwei Yang, Zuxin Liu, Juntao Tan, Prafulla~K Choubey, Tian Lan, Jason Wu, Huan Wang, and 1 others. 2024{\natexlab{b}}.
\newblock Agentlite: A lightweight library for building and advancing task-oriented llm agent system.
\newblock \emph{arXiv preprint arXiv:2402.15538}.

\bibitem[{Liu et~al.(2024{\natexlab{c}})Liu, Hoang, Zhang, Zhu, Lan, Tan, Yao, Liu, Feng, RN et~al.}]{liu2024apigen}
Zuxin Liu, Thai Hoang, Jianguo Zhang, Ming Zhu, Tian Lan, Juntao Tan, Weiran Yao, Zhiwei Liu, Yihao Feng, Rithesh RN, and 1 others. 2024{\natexlab{c}}.
\newblock Apigen: Automated pipeline for generating verifiable and diverse function-calling datasets.
\newblock \emph{Advances in Neural Information Processing Systems}, 37:54463--54482.

\bibitem[{Lumer et~al.(2025)Lumer, Lee, Kim, and Kim}]{lumer2025scalemcp}
Ed~Lumer, Chang Lee, Jinsu Kim, and Yeong-Dae Kim. 2025.
\newblock {ScaleMCP}: Dynamic and auto-synchronizing model context protocol tools for llm agents.
\newblock \emph{arXiv preprint arXiv:2505.06416}.

\bibitem[{Ma et~al.(2024{\natexlab{a}})Ma, Zhang, Lin, Shu, and Wang}]{ma2024agentboard}
Yitong Ma, Zeyu Zhang, Zepu Lin, Ke~Shu, and Chen Wang. 2024{\natexlab{a}}.
\newblock {AgentBoard}: An analytical evaluation board of multi-turn llm agents.
\newblock \emph{arXiv preprint arXiv:2401.13178}.
\newblock Accepted at NeurIPS 2024.

\bibitem[{Ma et~al.(2024{\natexlab{b}})Ma, Zhang, Liu, Zhang, Tan, Shu, Niebles, Heinecke, Wang, Xiong et~al.}]{ma2024taco}
Zixian Ma, Jianguo Zhang, Zhiwei Liu, Jieyu Zhang, Juntao Tan, Manli Shu, Juan~Carlos Niebles, Shelby Heinecke, Huan Wang, Caiming Xiong, and 1 others. 2024{\natexlab{b}}.
\newblock Taco: Learning multi-modal action models with synthetic chains-of-thought-and-action.
\newblock \emph{arXiv preprint arXiv:2412.05479}.

\bibitem[{OpenAI(2023)}]{openai2023gpt4}
OpenAI. 2023.
\newblock Gpt-4 technical report.
\newblock \url{https://arxiv.org/abs/2303.08774}.

\bibitem[{Ouyang et~al.(2022)Ouyang, Wu, Jiang, Almeida, Wainwright, Mishkin, Zhang, Agarwal, Slama, Ray et~al.}]{ouyang2022training}
Long Ouyang, Jeff Wu, Xu~Jiang, Diogo Almeida, Carroll Wainwright, Pamela Mishkin, Chong Zhang, Sandhini Agarwal, Katarina Slama, Alex Ray, and 1 others. 2022.
\newblock Training language models to follow instructions with human feedback.
\newblock In \emph{Advances in Neural Information Processing Systems}, volume~35, pages 27730--27744.

\bibitem[{Qin et~al.(2023)Qin, Cai, Liang, Zhang, Zhao, Lin, Yao, Deng, Li, Dong et~al.}]{qin2023toolbench}
Yujia Qin, Shi-Cheng Cai, Y-H Liang, Y~Zhang, X~Zhao, Y~Lin, Y-H Yao, X~Deng, Z-K Li, C~Dong, and 1 others. 2023.
\newblock Toolllm: Facilitating large language models to master 16000+ real-world apis.
\newblock \emph{arXiv preprint arXiv:2307.16789}.

\bibitem[{Roveda et~al.(2023)}]{crewai}
Joao Roveda and 1 others. 2023.
\newblock {CrewAI}.
\newblock \url{https://github.com/joaomdmoura/crewAI}.

\bibitem[{Srivastava et~al.(2022)Srivastava, Rastogi, Rao, Shoeb, Abid, Fisch, Brown, Santoro, Gupta, Garriga et~al.}]{srivastava2022beyond}
Aarohi Srivastava, Abhinav Rastogi, Abhishek Rao, Abu~Awal Shoeb, Abubakar Abid, Adam Fisch, Adam~R Brown, Adam Santoro, Aditya Gupta, Aitor Garriga, and 1 others. 2022.
\newblock Beyond the imitation game: Quantifying and extrapolating the capabilities of language models.
\newblock In \emph{International Conference on Machine Learning}, pages 20499--20689. PMLR.

\bibitem[{Tan et~al.(2025)Tan, Yang, Liu, Liu, Murthy, Awalgaonkar, Zhang, Yao, Zhu, Kokane, Savarese, Wang, Xiong, and Heinecke}]{tan2025personabenchevaluatingaimodels}
Juntao Tan, Liangwei Yang, Zuxin Liu, Zhiwei Liu, Rithesh Murthy, Tulika~Manoj Awalgaonkar, Jianguo Zhang, Weiran Yao, Ming Zhu, Shirley Kokane, Silvio Savarese, Huan Wang, Caiming Xiong, and Shelby Heinecke. 2025.
\newblock \href {https://arxiv.org/abs/2502.20616} {Personabench: Evaluating ai models on understanding personal information through accessing (synthetic) private user data}.
\newblock \emph{Preprint}, arXiv:2502.20616.

\bibitem[{Tang et~al.(2024)Tang, Shi, Chen, Zhao, Zhuang, Zhang, Chen, and Luo}]{tang2024matrix}
Shuke Tang, Zexuan Shi, Wenhai Chen, Zhaowei Zhao, Zirui Zhuang, Guanguan Zhang, Feng Chen, and Jie Luo. 2024.
\newblock {MATRIX}: A multi-agent reinforcement learning environment for text-based social interaction simulation.
\newblock \emph{arXiv preprint arXiv:2405.02705}.

\bibitem[{Taori et~al.(2023)Taori, Gulrajani, Zhang, Dubois, Li, Guestrin, Liang, and Hashimoto}]{taori2023alpaca}
Rohan Taori, Ishaan Gulrajani, Tianyi Zhang, Yann Dubois, Xuechen Li, Carlos Guestrin, Percy Liang, and Tatsunori~B Hashimoto. 2023.
\newblock Stanford alpaca: An instruction-following llama model.
\newblock \emph{GitHub repository}.

\bibitem[{Wang et~al.(2023)Wang, Kordi, Mishra, Liu, Smith, Khashabi, and Hajishirzi}]{wang2023selfinstruct}
Yizhong Wang, Yeganeh Kordi, Swaroop Mishra, Alisa Liu, Noah~A Smith, Daniel Khashabi, and Hannaneh Hajishirzi. 2023.
\newblock Self-instruct: Aligning language model with self generated instructions.
\newblock In \emph{Proceedings of the 61st Annual Meeting of the Association for Computational Linguistics (Volume 1: Long Papers)}, pages 13280--13299.

\bibitem[{White et~al.(2024)White, Dooley, Roberts, Shtedritski, Pochinkov, Ku, Jain, Jha, Ren, Sleigh et~al.}]{white2024livebench}
Colin White, Samuel Dooley, Manley Roberts, Arka Shtedritski, Chris Pochinkov, Shay Ku, Neel Jain, Siddharth Jha, Jiayi Ren, John Sleigh, and 1 others. 2024.
\newblock {LiveBench}: A challenging, contamination-free llm benchmark.
\newblock \emph{arXiv preprint arXiv:2406.19314}.

\bibitem[{Wu et~al.(2023)}]{autogen}
Qingyun Wu and 1 others. 2023.
\newblock {AutoGen}: Enabling next-gen llm applications via multi-agent conversation framework.
\newblock \url{https://github.com/microsoft/autogen}.

\bibitem[{Xie et~al.(2024)Xie, Chen, Gao, Hsieh, Yao, Cao, Jin, Gao, Li, Shen et~al.}]{xie2024osworld}
Tianbao Xie, Danyang Chen, Zhao Gao, Chun-Che Hsieh, Tao Yao, Hongjin Cao, Zetian Jin, Yunwei Gao, Zhenmei Li, Yifei Shen, and 1 others. 2024.
\newblock Osworld: Benchmarking multimodal agents for open-ended tasks in real computer environments.
\newblock \emph{arXiv preprint arXiv:2404.07972}.

\bibitem[{Xu et~al.(2023)Xu, Sun, Zheng, Geng, Zhao, Fung, Lin, Wu, Li, Jiang et~al.}]{xu2023wizardlm}
Can Xu, Qingfeng Sun, Kai Zheng, Xiubo Geng, Pu~Zhao, Jiazhan Fung, Yixin Lin, Xingxin Wu, Wenfeng Li, Weiming Jiang, and 1 others. 2023.
\newblock {WizardLM}: Empowering large language models to follow complex instructions.
\newblock \emph{arXiv preprint arXiv:2304.12244}.

\bibitem[{Yan et~al.(2025)Yan, Liu, Kim, Lee, and Yang}]{yan2025mcpworld}
Meng Yan, Ruihang Liu, Jinsu Kim, Chang Lee, and Diyi Yang. 2025.
\newblock {MCPWorld}: A unified benchmarking testbed for api, gui, and hybrid computer use agents.
\newblock \emph{arXiv preprint arXiv:2506.07672}.

\bibitem[{Yao et~al.(2022)Yao, Chen, Gu, R-K, Y-F, Le, and Song}]{yao2022webshop}
Shunyu Yao, Howard Chen, John Gu, K~R-K, C~Y-F, Q~Le, and D~Song. 2022.
\newblock Webshop: Towards scalable real-world web interaction with grounded language agents.
\newblock In \emph{Advances in Neural Information Processing Systems}.

\bibitem[{Zhang et~al.(2023{\natexlab{a}})Zhang, Zhang, Wu, Li, Lu, Lin, Wang, Yu, Yu, Sun et~al.}]{zhang2023surveyagents}
Cheng Zhang, Quan Zhang, Zhipu Wu, Jun-Yan Li, Wen-Juan Lu, Chang-Gen Lin, Sa-Hai Wang, Bin Yu, Philip~S. Yu, Li-Rong Sun, and 1 others. 2023{\natexlab{a}}.
\newblock A survey on large language model based autonomous agents.
\newblock \emph{arXiv preprint arXiv:2308.11432}.

\bibitem[{Zhang et~al.(2024{\natexlab{a}})Zhang, Zhu, Li, Wang, Zhao, and Zhang}]{zhang2024mmlupro}
Jia-Chen Zhang, Yitong Zhu, Zhuohao Li, Ge~Wang, He~Zhao, and Min-Ling Zhang. 2024{\natexlab{a}}.
\newblock {MMLU-Pro}: A more robust and challenging multi-task language understanding benchmark.
\newblock \emph{arXiv preprint arXiv:2406.01574}.
\newblock Accepted at NeurIPS 2024.

\bibitem[{Zhang et~al.(2025)Zhang, Hoang, Zhu, Liu, Wang, Awalgaonkar, Prabhakar, Chen, Yao, Liu et~al.}]{zhang2025actionstudio}
Jianguo Zhang, Thai Hoang, Ming Zhu, Zuxin Liu, Shiyu Wang, Tulika Awalgaonkar, Akshara Prabhakar, Haolin Chen, Weiran Yao, Zhiwei Liu, and 1 others. 2025.
\newblock Actionstudio: A lightweight framework for data and training of large action models.
\newblock \emph{arXiv preprint arXiv:2503.22673}.

\bibitem[{Zhang et~al.(2024{\natexlab{b}})Zhang, Lan, Zhu, Liu, Hoang, Kokane, Yao, Tan, Prabhakar, Chen et~al.}]{zhang2024xlam}
Jianguo Zhang, Tian Lan, Ming Zhu, Zuxin Liu, Thai Hoang, Shirley Kokane, Weiran Yao, Juntao Tan, Akshara Prabhakar, Haolin Chen, and 1 others. 2024{\natexlab{b}}.
\newblock xlam: A family of large action models to empower ai agent systems.
\newblock \emph{arXiv preprint arXiv:2409.03215}.

\bibitem[{Zhang et~al.(2023{\natexlab{b}})Zhang, Qian, Liu, Heinecke, Meng, Liu, Yu, Wang, Savarese, and Xiong}]{zhang2023dialogstudio}
Jianguo Zhang, Kun Qian, Zhiwei Liu, Shelby Heinecke, Rui Meng, Ye~Liu, Zhou Yu, Huan Wang, Silvio Savarese, and Caiming Xiong. 2023{\natexlab{b}}.
\newblock Dialogstudio: Towards richest and most diverse unified dataset collection for conversational ai.
\newblock \emph{arXiv preprint arXiv:2307.10172}.

\bibitem[{Zheng et~al.(2024)Zheng, Kou, Kumar, Ngo, Zhang, Wang, Li, and Liang}]{zheng2024helm}
Leon Zheng, Serena Kou, Neel Kumar, Hieu Ngo, Boxin Zhang, Zhaohui Wang, Percy Li, and Percy Liang. 2024.
\newblock {HELM} {S}afety: Towards standardized safety evaluations of language models.
\newblock \emph{arXiv preprint arXiv:2405.09340}.

\bibitem[{Zheng et~al.(2023)Zheng, Chiang, Sheng, Zhuang, Wu, Zhuang, Lin, Li, Brooks, Xing et~al.}]{zheng2023mtbench}
Lianmin Zheng, Wei-Lin Chiang, Ying Sheng, Siyuan Zhuang, Zhanghao Wu, Yonghao Zhuang, Zi~Lin, Zhuohan Li, Dacheng Brooks, Eric Xing, and 1 others. 2023.
\newblock Judging {LLM-as-a-judge} with {MT-Bench} and {Chatbot Arena}.
\newblock \emph{arXiv preprint arXiv:2306.05685}.

\bibitem[{Zhou et~al.(2023)Zhou, Xu, Zhu, Zhou, Li, Li, Liu, Gu, Zhang, Li et~al.}]{zhou2023webarena}
Shuyan Zhou, Frank~F Xu, Hao Zhu, Xuhui Zhou, Robert Li, Zhengyun Li, C~Liu, P~Gu, Y~Zhang, C~Li, and 1 others. 2023.
\newblock {WebArena}: A realistic web environment for building autonomous agents.
\newblock \emph{arXiv preprint arXiv:2307.13854}.

\bibitem[{Zhu et~al.(2024)Zhu, Lin, Gao, Zhou, Zhang, Liu, Zheng, Li, Wang, Wang et~al.}]{zhu2024survey}
Jihan Zhu, Zhizheng Lin, Jun Gao, Zhaoxuan Zhou, Yaodong Zhang, Zhaofan Liu, Ceyao Zheng, Cheng Li, Zhaokai Wang, Zili Wang, and 1 others. 2024.
\newblock A survey of ai agent evaluation: A hundred unsolved problems and a one-stop open-source library.
\newblock \emph{arXiv preprint arXiv:2406.09844}.

\end{thebibliography}

% \bibliography{reference}

\newpage
\appendix
\onecolumn
\section{Comprehensive Experiments and Results}
\label{appendix:results}

We present a comprehensive evaluation of 10 state-of-the-art LLM models across 5 diverse domains using the \oursystemname framework. Our analysis encompasses 5k trajectory records and 5k completion records with detailed individual task analysis, plus 50 model-domain combinations from systematic LLM judger evaluation, representing the most extensive evaluation of LLM tool-use capabilities to date.

\subsection{Experimental Setup}
\label{subsec:setup}

\paragraph{Model Selection}
Our evaluation includes 10 models spanning different architectures and capabilities:
\begin{itemize}
    \item {OpenAI Models} (7): GPT-4o, GPT-4o-mini, GPT-4.1-mini, GPT-4.1-nano, O3, O3-mini, O4-mini
    \vspace{-5pt}
    \item {Open-Source Models} (3): Mistral-Small-24B, Qwen3-32B, Qwen3-30B-A3B
\end{itemize}

\paragraph{Domain and Tools Coverage}
We evaluate performance across 5 diverse application domains with the following tools:
\begin{itemize}
    \item {Finance}: Stock prices, financial data, market analysis, portfolio management
    \item {Healthcare}: Medical terminology lookup, drug information, clinical trials, health topics, PubMed search
    \vspace{-5pt}
    \item {Airbnb}:\texttt{airbnb\_search}, \texttt{airbnb\_listing\_details}
    \vspace{-5pt}
    \item {Sports}: Team statistics, player information, game schedules, league standings  
    \vspace{-5pt}
    \item {National Parks}: Park information, visitor services, trail details, facility booking
    \vspace{-5pt}
    
\end{itemize}

\section{Evaluation Criteria}\label{appendix:evaluation criteria}

To holistically assess the capabilities of AI agents in complex task environments, we adopt a two-dimensional evaluation criteria: (1) \textbf{Tool Call Performance}, which measures the correctness of predicted tool usage against the ground truth, and (2) \textbf{LLM Judger Performance}, which scores the overall quality of execution trajectories and task completion using rubric-based judgment. 
These complementary dimensions enable both precise operational assessment and high-level behavioral evaluation.

\subsection{Tool Call Criteria}

Tool call evaluation is conducted using the MCP Model Evaluator’s \texttt{analyze} command. This system compares the agent's predicted tool usage with the ground truth across both \textit{strict} and \textit{flexible} matching protocols. 

\paragraph{Strict Matching:} Requires exact correspondence on tool names, parameter values, and execution order. It represents a binary success paradigm: either the task is fully correct or it fails.

\paragraph{Flexible Matching:} Allows partial credit by applying similarity thresholds for parameter values ($\geq$ 0.6) and tool order ($\geq$ 0.5). It reflects more tolerant criteria aligned with approximate but contextually appropriate predictions.

\paragraph{Metric Definitions:}
\begin{itemize}
    \item \textbf{Average Name Match Score}: Measures the proportion of correctly predicted tool names.
    \item \textbf{Average Parameter Match Score}: Assesses correctness or similarity of parameter values.
    \item \textbf{Average Order Match Score}: Evaluates sequence alignment between predicted and actual tool calls.
    \item \textbf{Average Overall Score}: A weighted combination of the above metrics, using configurable weights (default: 0.4 for name and parameter, 0.2 for order).
\end{itemize}

In addition to aggregate statistics, the system also provides fine-grained diagnostics such as missing/extra tools, parameter mismatches, and tool-specific success rates, facilitating detailed error analysis.

\subsection{LLM Judger Criteria}\label{appendix:llm-judger criteria}

Beyond tool-level correctness, we incorporate a rubric-based judgment system to evaluate execution quality from a human-centric perspective. This LLM Judger scores each trajectory on multiple aspects of agent behavior, encompassing planning, execution logic, tool use, and final task outcomes.

\paragraph{Trajectory Evaluation Aspects:}
\begin{itemize}
    \item \textbf{Planning:} Measures task understanding and decomposition.
    \item \textbf{Execution Flow:} Assesses the logical coherence and sequencing of actions.
    \item \textbf{Tool Selection \& Usage:} Evaluates appropriateness of tools and correctness of parameters.
    \item \textbf{Adaptability:} Captures the agent's response to unexpected results and changing context.
    \item \textbf{Efficiency:} Reflects the conciseness and lack of redundancy in the solution.
    \item \textbf{Context Awareness:} Checks whether the agent maintains relevant constraints and state.
\end{itemize}

\paragraph{Task Completion Evaluation Aspects:}
\begin{itemize}
    \item \textbf{Requirement Coverage:} Degree to which the output fulfills all task goals.
    \item \textbf{Accuracy:} Factual and logical correctness of the output.
    \item \textbf{Completeness:} Depth and breadth of the agent’s response.
    \item \textbf{Usefulness:} Practical value and user relevance of the solution.
\end{itemize}

Each aspect is scored on a continuous scale from 0.0 to 1.0, allowing nuanced grading. A high-quality trajectory demonstrates coherent planning, accurate tool use, robust adaptability, and a final response that is complete, accurate, and valuable.
\section{Model Versions}

Table \ref{appendix-model-versions} presents the model names and their corresponding versions used in this work. For certain models, including the o3 and o4 series, the default temperature was used, as their APIs do not support temperature adjustment. We set the temperature to a constant value of 0.01 for the rest  to ensure reproducible results.

\begin{table*}[t]\small
\centering
\caption{Model names and their corresponding versions.}
\vspace{-5pt}
\label{tab:tool_accuracy}
\begin{tabular}{ll}
\toprule
\textbf{Model} & \textbf{Model Version}  \\
\midrule
mistral-small-24b & mistralai/Mistral-Small-3.2-24B-Instruct-2506 \\
qwen3-30b-a3b & Qwen/Qwen3-30B-A3B \\
qwen3-32b & Qwen/Qwen3-32B \\
gpt4.1-mini & gpt-4.1-mini-2025-04-14 \\
gpt4.1-nano & gpt-4.1-nano-2025-04-14 \\
gpt-4o-mini & gpt-4o-mini-2024-07-18 \\
gpt-4o & gpt-4o-2024-08-06 \\
gpt4.1 & gpt-4.1-2025-04-14 \\
o3-mini & o3-mini-2025-01-31 \\
o4-mini & o4-mini-2025-04-16 \\
o3 & o3-2025-04-16 \\
\bottomrule
\end{tabular}\label{appendix-model-versions}
\end{table*}

\subsection{Overall Performance Results}
\label{subsec:overall}

\subsubsection{Performance Statistics}

\begin{table}[t]\small
\centering
\caption{Overall performance statistics.}
\vspace{-5pt}
\label{tab:overall_stats}
\begin{tabular}{lcc}
\toprule
\textbf{Metric} & \textbf{Individual Tasks} & \textbf{LLM Judger} \\
\midrule
Total Evaluations & 10,115 & 50 \\
Average Trajectory Score & 0.839 ± 0.071 & 0.779 ± 0.084 \\
Average Completion Score & 0.774 ± 0.090 & 0.679 ± 0.217 \\
Performance Gap & 0.065 & 0.100 \\
Success Rate & 92.7\% & 93.9\% ± 14.3\% \\
\bottomrule
\end{tabular}
\end{table}

Table~\ref{tab:overall_stats} presents the comprehensive performance statistics across our multi-level evaluation. The substantial difference in evaluation scale, 10,115 individual tasks versus 50 model-domain combinations, demonstrates the complementary nature of our multi-level approach. Individual task analysis provides granular insights into specific failure modes and success patterns, while LLM judger evaluation offers systematic assessment of overall performance quality.
Our results reveal a consistent \textit{performance gap} of 0.065-0.100 between trajectory execution and completion quality across both evaluation methodologies. This indicates that while models excel at planning and executing tool usage, they face challenges in producing high-quality final outputs.

The trajectory performance scores consistently exceed completion scores, confirming the universal execution-completion gap across evaluation methodologies. This  gap represents a fundamental limitation in current architectures, where models demonstrate superior procedural reasoning compared to output synthesis capabilities.

Notably, the success rates indicate that models can complete most tasks without catastrophic failures, but the quality of outputs varies significantly. The higher standard deviation in completion scores suggests greater variability in output quality compared to execution consistency, indicating that completion quality is more sensitive to task complexity and model capabilities.

The convergent findings across both evaluation methodologies strengthen the reliability of our results and confirm that the trajectory-completion gap represents a genuine architectural challenge rather than an evaluation artifact.

\subsubsection{Model Performance Hierarchy}
\begin{figure}[t]
\centering
\includegraphics[width=\textwidth]{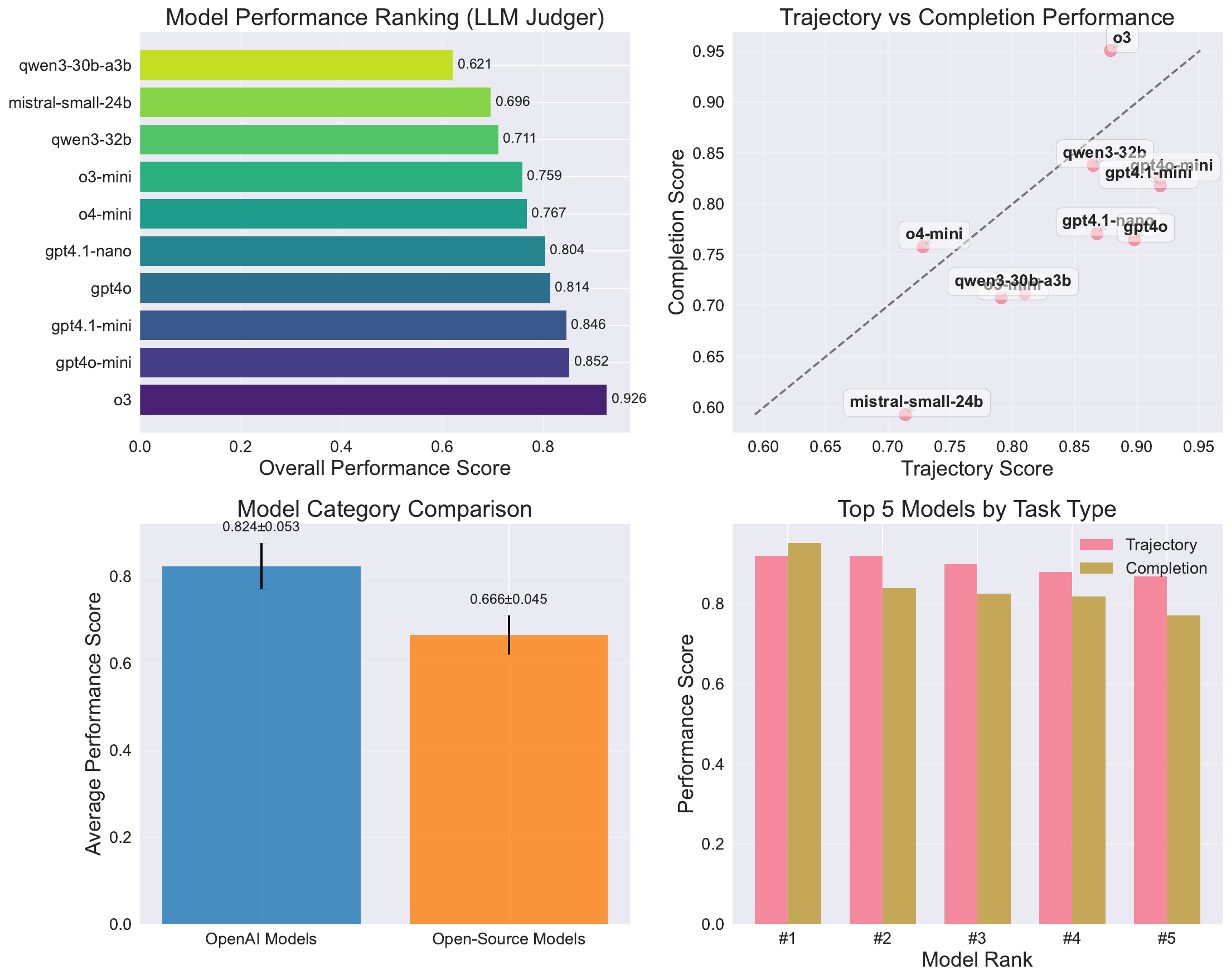}
\caption{Model performance analysis: (a) Overall ranking by LLM judger, (b) Trajectory vs completion performance scatter plot, (c) Model category comparison, (d) Top 5 models by task type.}
\label{fig:model_performance}
\end{figure}

Figure~\ref{fig:model_performance} illustrates the comprehensive model performance analysis across different evaluation dimensions, revealing distinct performance patterns and architectural characteristics.

Figure~\ref{fig:model_performance}(a) demonstrates the overall performance ranking based on LLM judger evaluation, with O3 leading at 0.926, followed closely by GPT-4o-mini (0.852) and GPT-4.1-mini (0.846). The ranking reveals a clear performance hierarchy, with OpenAI models dominating the top positions and open-source models showing more variable performance. Notably, the performance distribution shows a gradual decline from top performers (>0.8) to lower-performing models (>0.6), indicating significant capability differences across model architectures.

Figure~\ref{fig:model_performance}(b) presents the trajectory vs completion performance scatter plot, revealing the fundamental execution-completion trade-off. Most models cluster above the diagonal line, indicating superior trajectory execution compared to completion quality. O3 stands out as the model below the diagonal, demonstrating exceptional completion quality relative to its trajectory performance. The tight clustering of GPT models (GPT-4o, GPT-4o-mini, GPT-4.1-mini) in the upper-right region indicates consistent high performance across both dimensions, while open-source models show greater variance and generally lower performance.

Figure~\ref{fig:model_performance}(c) illustrates the stark performance gap between model categories, with OpenAI models achieving an average score of 0.824 compared to 0.676 for open-source models, showing a significant performance difference. This performance gap highlights the current disparity in tool-use capabilities between proprietary and open-source architectures. The error bars indicate that while OpenAI models show consistent performance with low variance, open-source models exhibit higher variance, suggesting less stable tool-use capabilities.

Figure~\ref{fig:model_performance}(d) compares the top 5 models by task type, revealing specialization patterns across trajectory execution and completion quality. GPT-4.1-mini, GPT-4o-mini, and GPT-4o dominate trajectory performance with scores above 0.9, while O3, Qwen3-32B, and GPT-4o-mini lead in completion quality. This analysis highlights that model excellence in one dimension doesn't guarantee equivalent performance in the other, suggesting different optimization strategies across model architectures.

\begin{table}[t]\small
\centering
\caption{Model performance rankings.}
\vspace{-5pt}
\label{tab:model_ranking}
\begin{tabular}{lccc}
\toprule
\textbf{Model} & \textbf{Trajectory} & \textbf{Completion} & \textbf{Overall (LLM)} \\
\midrule
Mistral-Small-24B & 0.714 ± 0.253 & 0.593 ± 0.282 & 0.719 \\

Qwen3-30B-A3B & 0.810 ± 0.251 & 0.712 ± 0.272 & 0.609 \\
Qwen3-32B & 0.865 ± 0.185 & 0.838 ± 0.216 & 0.705 \\
GPT-4.1-mini & \textbf{0.915} ± 0.098 & 0.805 ± 0.143 & 0.839 \\
GPT-4o-mini & 0.911 ± 0.113 & 0.810 ± 0.164 & 0.839 \\
GPT-4o & 0.910 ± 0.112 & 0.806 ± 0.157 & 0.806 \\
GPT-4.1-nano & 0.866 ± 0.108 & 0.757 ± 0.164 & 0.796 \\
O3-mini & 0.760 ± 0.289 & 0.669 ± 0.266 & 0.748 \\
O4-mini & 0.698 ± 0.352 & 0.742 ± 0.363 & 0.755 \\
O3 & 0.857 ± 0.179 & \textbf{0.951} ± 0.171 & \textbf{0.926} \\
\bottomrule
\end{tabular}
\end{table}

The detailed analysis in Table~\ref{tab:model_ranking} reveals distinct performance patterns that reflect fundamental architectural differences across model families. The trajectory execution rankings demonstrate remarkable consistency among top-tier OpenAI models, with GPT-4.1-mini, GPT-4o-mini, and GPT-4o achieving nearly identical scores within a 0.005 range. This tight clustering suggests that these models share similar procedural reasoning capabilities, likely stemming from comparable training methodologies and architectural foundations.

The completion quality rankings present a more diverse landscape, with O3 achieving exceptional performance that surpasses all other models by at least 0.113 points. This dominance in completion quality, combined with O3's moderate trajectory performance, indicates a fundamentally different optimization strategy, prioritizing output synthesis over execution consistency. Qwen3-32B's strong completion performance demonstrates that open-source models can achieve competitive output quality, though this doesn't translate to overall performance leadership due to lower trajectory scores.
The standard deviation patterns reveal important insights about model reliability. GPT-4.1-mini shows the lowest trajectory variance, indicating exceptional consistency in execution quality. 

The performance tiers evident in the rankings, top performers (>0.8), mid-tier models (0.7-0.8), and lower performers (0.6-0.7), reflect fundamental capability differences rather than minor variations. The performance gap between different models represents a substantial difference in practical tool-use effectiveness, highlighting the current disparity in LLM capabilities for complex tool-use scenarios.

Through Table~\ref{tab:model_ranking}, we can summarize that: 
\vspace{-5pt}
\begin{itemize}
    \item \textit{Trajectory Leaders}: GPT-4.1-mini, GPT-4o-mini, GPT-4o demonstrate superior execution consistency.
    \vspace{-5pt}
    \item \textit{Completion Leaders}: O3, Qwen3-32B, GPT-4o-mini excel in output quality.
    \vspace{-5pt}
    \item \textit{Overall Champions}: O3 achieves the best combined performance, followed by GPT-4o-mini and GPT-4.1-mini.
    \vspace{-5pt}
    \item \textit{Category Gap}: OpenAI models (0.824 avg) significantly outperform open-source models (0.676 avg) by 0.148 points.
\end{itemize}

\subsection{Domain Performance Analysis}
\label{subsec:domains}

\begin{figure}[t]
\centering
\includegraphics[width=\textwidth]{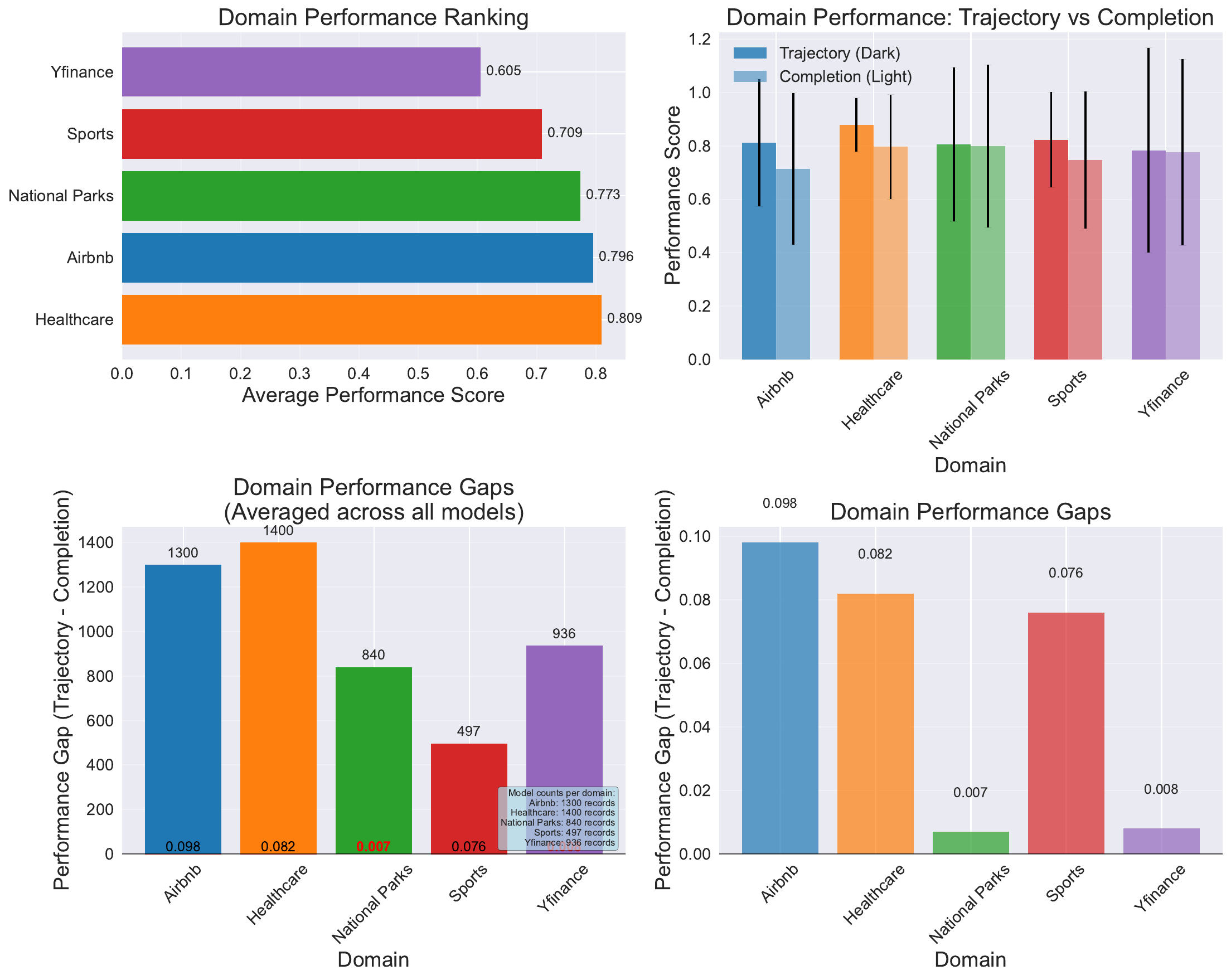}
\caption{Domain performance analysis: (a) Domain ranking by LLM judger, (b) Trajectory vs completion comparison, (c) Task distribution, (d) Performance gaps by domain.}
\label{fig:domain_analysis}
\end{figure}

Figure~\ref{fig:domain_analysis} presents comprehensive domain performance analysis across multiple dimensions, revealing significant variations in task complexity and API design quality across different application domains.

Figure~\ref{fig:domain_analysis}(a) establishes the domain performance hierarchy through LLM judger evaluation, with Healthcare leading at 0.809, followed by Airbnb (0.796), National Parks (0.773), Sports (0.709), and Finance (0.605). This ranking reflects the varying complexity of domain-specific tool ecosystems and the structural quality of underlying APIs. Healthcare's dominance suggests that well-standardized medical terminologies and structured data formats facilitate superior LLM performance, while Finance's challenges indicate the complexity of financial data interpretation and market volatility.

Figure~\ref{fig:domain_analysis}(b) compares trajectory and completion performance across domains, revealing domain-specific execution-completion patterns. Healthcare demonstrates the highest trajectory performance with strong completion quality, indicating well-aligned API responses with user expectations. National Parks shows balanced performance with the smallest gap, suggesting that park information APIs provide clear, structured outputs. Finance exhibits lower trajectory performance, reflecting the complexity of financial data interpretation, while Airbnb shows a significant execution-completion gap, indicating challenges in translating property searches into comprehensive recommendations.

Figure~\ref{fig:domain_analysis}(c) illustrates the task distribution across domains, with Healthcare (1,400 tasks) and Airbnb (1,300 tasks) representing the largest evaluation sets, followed by Finance (936 tasks), National Parks (840 tasks), and Sports (497 tasks). This distribution reflects both the complexity and practical importance of these domains in real-world applications. The substantial task counts in Healthcare and Airbnb provide robust statistical power for performance analysis, while the smaller Sports dataset still offers sufficient coverage for meaningful evaluation.

Figure~\ref{fig:domain_analysis}(d) reveals domain-specific performance gaps, highlighting fundamental differences in task complexity and API design quality. Airbnb exhibits the largest gap (0.098), suggesting that while models can effectively search for properties, they struggle to synthesize results into useful, comprehensive recommendations. Healthcare and Sports show moderate gaps (0.082 and 0.076), indicating balanced but execution-favored performance. National Parks demonstrates the smallest gap (0.007), reflecting the alignment between structured park information and model output capabilities. Finance shows a small gap (0.008), suggesting that financial data APIs provide clear outputs that match user requirements.

\begin{table}[t]\small
\centering
\caption{Domain performance characteristics.}
\vspace{-5pt}
\label{tab:domain_performance}
\begin{tabular}{lcccc}
\toprule
\textbf{Domain} & \textbf{Trajectory} & \textbf{Completion} & \textbf{Gap} & \textbf{Tasks} \\
\midrule
Healthcare & \textbf{0.879} ± 0.101 & \textbf{0.797} ± 0.196 & 0.082 & 1,400 \\
Airbnb & 0.812 ± 0.239 & 0.714 ± 0.285 & 0.098 & 1,300 \\
Sports & 0.823 ± 0.179 & 0.747 ± 0.257 & 0.076 & 497 \\
Finance & 0.784 ± 0.359 & 0.776 ± 0.329 & 0.008 & 936 \\
National Parks & 0.806 ± 0.184 & 0.799 ± 0.187 & 0.007 & 840 \\
\bottomrule
\end{tabular}
\end{table}

The comprehensive domain analysis in Table~\ref{tab:domain_performance} reveals fundamental differences in task complexity and API ecosystem quality across application domains. Healthcare's dominance in both trajectory and completion performance reflects the mature state of medical informatics infrastructure, where standardized terminologies (ICD-10, SNOMED-CT) and well-documented APIs facilitate effective tool use. The moderate gap indicates that medical APIs generally produce outputs that align well with user expectations, though synthesis challenges remain.

The variance patterns provide crucial insights into domain stability and predictability. Healthcare shows the lowest trajectory variance (±0.101), indicating consistent performance across diverse medical tasks, while Finance exhibits the highest variance, reflecting the inherent volatility and complexity of financial data interpretation. This high variance in Finance, combined with its strong completion performance, suggests that while financial APIs can produce high-quality outputs, the execution path varies significantly based on market conditions and data availability.

Task distribution analysis reveals the practical importance and complexity of each domain. Healthcare (1,400 tasks) and Airbnb (1,300 tasks) represent the most comprehensive evaluations, providing robust statistical power for performance analysis. The substantial task counts in these domains reflect both their real-world importance and the complexity of their tool ecosystems. Sports' dataset (497 tasks) still provides sufficient coverage for meaningful analysis, while Finance (936 tasks) and National Parks (840 tasks) offer balanced evaluation sets.

The performance gap analysis reveals domain-specific challenges in translating execution success into output quality. Airbnb's large gap indicates fundamental challenges in property recommendation synthesis—while models can effectively search and filter properties, they struggle to present results in comprehensive, actionable formats. Conversely, National Parks' minimal gap suggests that park information APIs are well-designed for direct consumption, requiring minimal synthesis for user value.

The domain ranking hierarchy reflects not just API quality but also the inherent complexity of data interpretation and user requirements. Finance's challenges stem from the need to interpret complex financial data, market trends, and provide actionable investment insights. This complexity manifests in lower execution consistency and moderate completion quality, highlighting the challenges of real-world financial data analysis.

Summarizing Table~\ref{tab:domain_performance}, we can find that:  
\begin{itemize}
    \item \textit{Healthcare} emerges as the most successful domain, likely due to well-structured medical APIs and standardized terminology.
    \vspace{-5pt}
    \item \textit{National Parks} shows the smallest performance gap, indicating better alignment between execution and output quality.
    \vspace{-5pt}
    \item \textit{Finance} proves most challenging, reflecting complex financial data interpretation requirements.
    \vspace{-5pt}
    \item \textit{Airbnb} demonstrates the largest performance gap, suggesting execution-completion misalignment issues
\end{itemize}

\subsection{Aspect-Level Performance Analysis}
\label{subsec:aspects}

\begin{figure}[t]
\centering
\includegraphics[width=\textwidth]{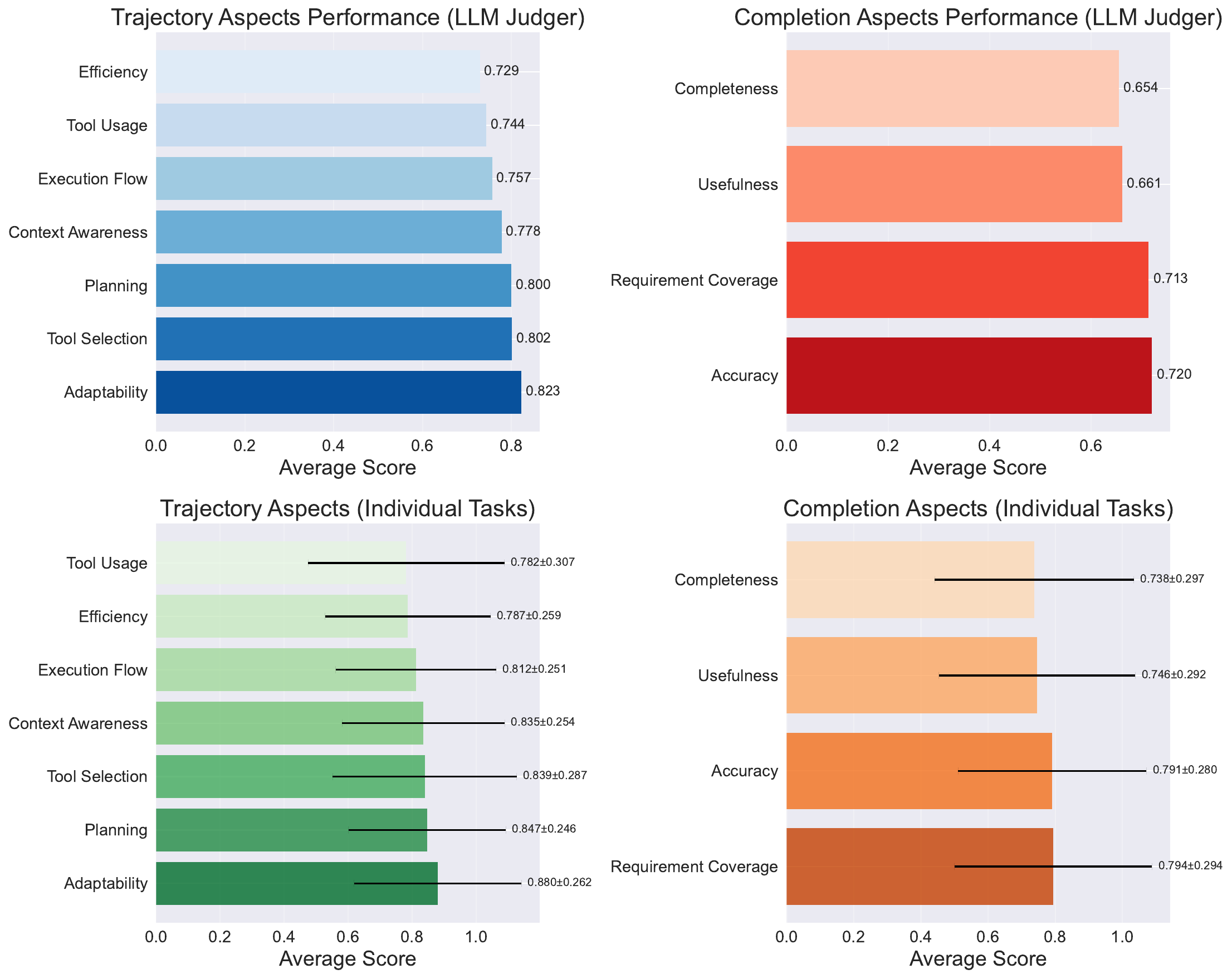}
\caption{Aspect performance analysis: (a) Trajectory aspects (LLM judger), (b) Completion aspects (LLM judger), (c) Trajectory aspects (individual tasks), (d) Completion aspects (individual tasks).}
\label{fig:aspect_analysis}
\end{figure}

Figure~\ref{fig:aspect_analysis} provides detailed analysis of trajectory and completion aspects across both evaluation methodologies, revealing the specific strengths and weaknesses of current LLM architectures in tool-use scenarios.

Figure~\ref{fig:aspect_analysis}(a) presents trajectory aspect performance from LLM judger evaluation, revealing models' strategic and execution capabilities. Planning emerges as the strongest aspect, indicating that models excel at formulating coherent action sequences for complex tasks. Adaptability and Tool Selection follow closely, demonstrating models' ability to adjust strategies and choose appropriate tools. However, Tool Usage represents the primary bottleneck, suggesting that while models can select correct tools, they struggle with precise parameter specification and effective tool utilization.

Figure~\ref{fig:aspect_analysis}(b) illustrates completion aspect performance, highlighting the quality challenges in LLM output generation. Requirement Coverage leads among completion aspects, indicating that models generally address the core requirements of given tasks. Accuracy closely follows, suggesting that when models produce outputs, they tend to be factually correct. However, Completeness and Usefulness  lag significantly, revealing that models often produce incomplete or less practical outputs despite correct execution.

Figure~\ref{fig:aspect_analysis}(c) shows trajectory aspects from individual task evaluation, providing granular insights into execution quality. Adaptability and Planning achieve the highest scores, confirming models' strategic strengths across evaluation methodologies. Tool Selection (and Context Awareness demonstrate solid performance, while Execution Flow and Efficiency (0.789) show moderate scores. Tool Usage again emerges as the weakest aspect, consistent with LLM judger findings and confirming this as a fundamental limitation.

Figure~\ref{fig:aspect_analysis}(d) presents completion aspects from individual task evaluation, corroborating the completion quality challenges identified in LLM judger analysis. Requirement Coverage and Accuracy lead the completion aspects, indicating that models generally produce relevant and correct outputs. However, Usefulness and Completeness show lower scores, suggesting that while outputs are accurate, they often lack practical value or comprehensive coverage of user needs.

\subsubsection{Trajectory Aspects}

\begin{table}[t]\small
\centering
\caption{Trajectory aspect performance.}
\vspace{-5pt}
\label{tab:trajectory_aspects}
\begin{tabular}{lcc}
\toprule
\textbf{Aspect} & \textbf{Individual Tasks} & \textbf{LLM Judger} \\
\midrule
Adaptability & \textbf{0.874} ± 0.245 & 0.810 ± 0.163 \\
Planning & 0.857 ± 0.226 & \textbf{0.818} ± 0.163 \\
Tool Selection & 0.842 ± 0.272 & 0.805 ± 0.163 \\
Context Awareness & 0.839 ± 0.238 & 0.786 ± 0.163 \\
Execution Flow & 0.818 ± 0.235 & 0.762 ± 0.163 \\
Efficiency & 0.789 ± 0.246 & 0.726 ± 0.163 \\
Tool Usage & 0.776 ± 0.296 & \textbf{0.722} ± 0.163 \\
\bottomrule
\end{tabular}
\end{table}

The trajectory aspect analysis in Table~\ref{tab:trajectory_aspects} reveals the hierarchical structure of LLM tool-use capabilities, with clear distinctions between strategic and operational competencies. The consistent ranking across both evaluation methodologies validates the reliability of our assessment framework and confirms fundamental patterns in current LLM architectures.

Strategic aspects, like Adaptability, Planning, and Tool Selection, demonstrate superior performance, indicating that models excel at high-level reasoning tasks. The strong adaptability scores suggest that models can effectively adjust their strategies when initial approaches fail, while excellent planning capabilities indicate sophisticated multi-step reasoning. These strategic strengths reflect the success of current training methodologies in developing abstract reasoning capabilities.

Operational aspects show more variable performance, with Context Awareness performing well but Execution Flow and Efficiency showing moderate scores. The decline in operational performance suggests that while models can formulate good strategies, they face challenges in optimal execution. This pattern indicates that current architectures may be optimized for reasoning rather than execution efficiency.

Tool Usage emerges as the critical bottleneck, representing the lowest-performing aspect across both evaluation methodologies. This fundamental limitation reflects challenges in precise parameter specification, API interaction patterns, and error handling. The high variance in Tool Usage indicates significant inconsistency across different tools and domains, suggesting that current models lack robust tool interaction capabilities.

The variance patterns reveal important insights about aspect stability. Planning shows relatively low variance, indicating consistent strategic capabilities, while Tool Selection exhibits higher variance, suggesting that tool choice quality varies significantly based on task complexity and domain characteristics. This pattern highlights the need for improved tool interaction training and more robust parameter specification mechanisms.

\subsubsection{Completion Aspects}

\begin{table}[t]\small
\centering
\caption{Completion aspect performance.}
\vspace{-5pt}
\label{tab:completion_aspects}
\begin{tabular}{lcc}
\toprule
\textbf{Aspect} & \textbf{Individual Tasks} & \textbf{LLM Judger} \\
\midrule
Requirement Coverage & \textbf{0.793} ± 0.278 & \textbf{0.709} ± 0.217 \\
Accuracy & 0.785 ± 0.263 & 0.706 ± 0.217 \\
Usefulness & 0.741 ± 0.276 & 0.654 ± 0.217 \\
Completeness & 0.730 ± 0.282 & 0.645 ± 0.217 \\
\bottomrule
\end{tabular}
\end{table}

The completion aspect analysis in Table~\ref{tab:completion_aspects} reveals a fundamental dichotomy in LLM output capabilities, with models demonstrating strength in correctness but weakness in synthesis quality. The consistent ranking across both evaluation methodologies, where Requirement Coverage > Accuracy > Usefulness > Completeness, indicates systematic patterns in how current models generate outputs.

Requirement Coverage and Accuracy represent the strongest completion aspects, indicating that models generally understand task requirements and produce factually correct outputs. The close performance between these aspects (0.008 difference in individual tasks) suggests that when models address requirements, they typically do so accurately. This strength reflects the success of current training methodologies in developing factual consistency and requirement understanding.

However, the significant drop in Usefulness and Completeness reveals fundamental limitations in output synthesis capabilities. The gap between Accuracy and Usefulness indicates that while outputs are correct, they often lack practical value for users. This pattern suggests that models struggle to transform correct information into actionable insights, highlighting a critical limitation in current architectures.

The variance patterns provide additional insights into completion quality stability. Requirement Coverage shows high variance, indicating significant inconsistency in how well models address different types of requirements. Completeness exhibits the highest variance, suggesting that output comprehensiveness varies dramatically based on task complexity and domain characteristics. This instability in completion quality contrasts with the more consistent trajectory performance, indicating that output generation is more sensitive to task and model variations.

The systematic decline from correctness (Accuracy: 0.785) to synthesis quality (Completeness: 0.730) represents a 0.055-point degradation, highlighting the fundamental challenge in current LLM architectures. Models excel at retrieving and presenting correct information but struggle to synthesize comprehensive, useful outputs. This pattern suggests that future development should prioritize output synthesis capabilities, particularly in generating complete and practically useful responses.

Through Table~\ref{tab:completion_aspects}, we can get some interesting findings:
\vspace{-5pt}
\begin{itemize}
    \item \textit{Strongest Trajectory Aspects}: Adaptability and Planning demonstrate models' strategic capabilities.
    \vspace{-5pt}
    \item \textit{Weakest Trajectory Aspect}: Tool Usage represents the primary bottleneck in current LLM tool-use capabilities.
    \vspace{-5pt}
    \item \textit{Completion Quality Leaders}: Requirement Coverage and Accuracy  show models generally address task requirements correctly.
    \vspace{-5pt}
    \item \textit{Completion Challenges}: Completeness and Usefulness indicate room for improvement in output quality.
\end{itemize}

\subsection{Performance Gap Analysis}
\label{subsec:gaps}

\begin{figure}[t]
\centering
\includegraphics[width=0.9\linewidth]{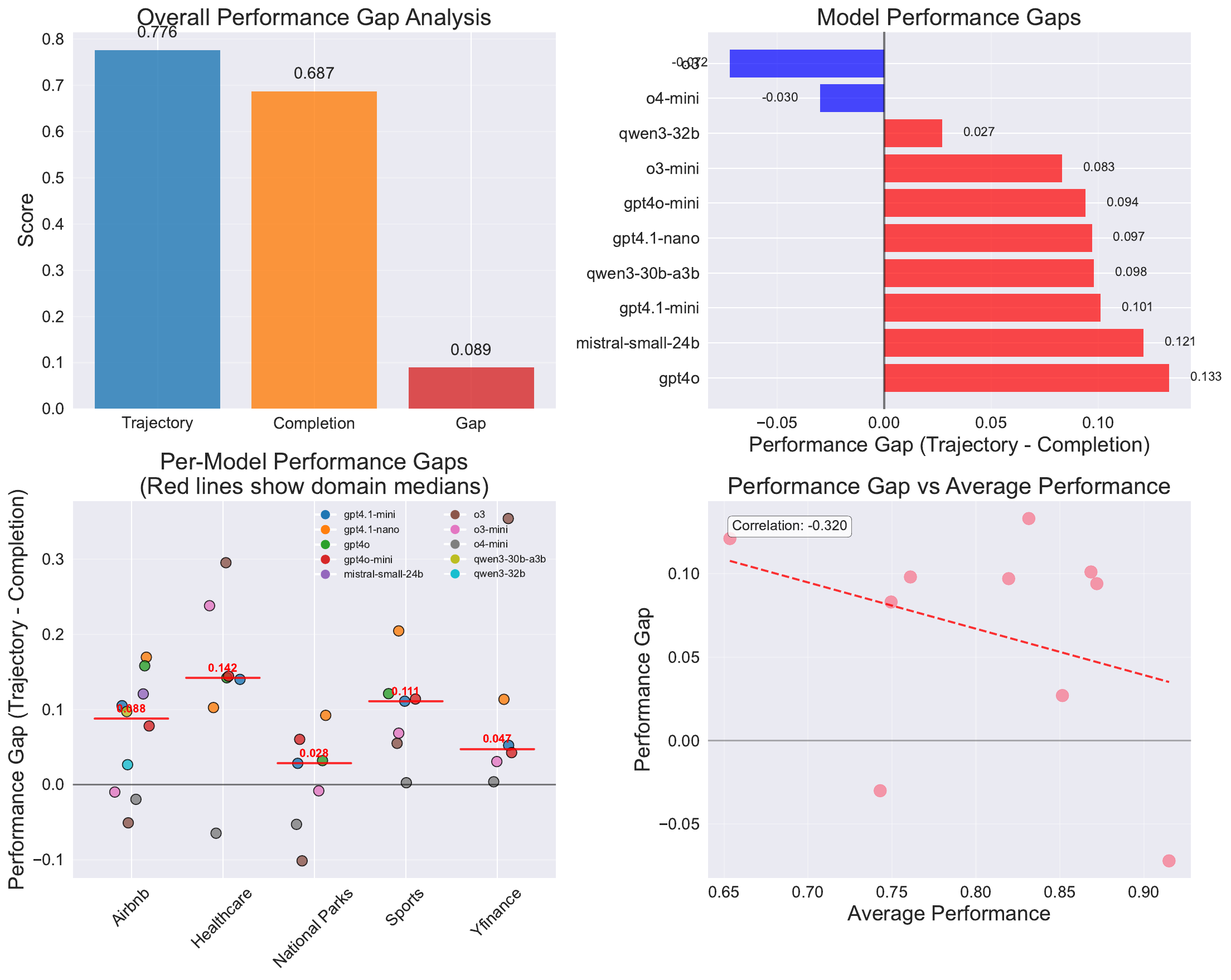}
\caption{Performance gap analysis: (a) Overall gap distribution, (b) Model-wise gaps, (c) Domain-wise gaps, (d) Gap-performance correlation}
\label{fig:gap_analysis}
\vspace{-10pt}
\end{figure}

Figure~\ref{fig:gap_analysis} presents comprehensive analysis of performance gaps across models and domains, revealing fundamental patterns in LLM tool-use capabilities and architectural characteristics.

Figure~\ref{fig:gap_analysis}(a) demonstrates the overall performance gap distribution, showing trajectory execution consistently outperforming completion quality across all evaluations. This universal phenomenon indicates that while models excel at procedural reasoning and tool orchestration, they struggle with synthesis and output generation. The distribution reveals that execution capabilities are more mature than completion quality in current LLM architectures, suggesting optimization priorities in model development.

Figure~\ref{fig:gap_analysis}(b) illustrates model-specific gap patterns, revealing distinct architectural characteristics across different LLM families. Most models show positive gaps ranging from 0.027 (Qwen3-32B) to 0.133 (GPT-4o), indicating execution-favored performance. Remarkably, O3 exhibits a negative gap (-0.072), demonstrating superior completion quality relative to trajectory execution, which is a unique characteristic that contributes to its overall performance leadership. The OpenAI models (GPT-4.1-mini, GPT-4o-mini, GPT-4o) demonstrate moderate gaps, suggesting balanced but execution-favored architectures, while open-source models show larger gaps, indicating architectural limitations in output synthesis.

Figure~\ref{fig:gap_analysis}(c) reveals domain-specific performance characteristics that reflect task complexity and API design quality. All domains exhibit positive gaps, but with significant variation ranging from 0.007 (National Parks) to 0.098 (Airbnb). National Parks demonstrates the smallest gap, suggesting that park information APIs provide clear, structured outputs that align well with model execution capabilities. In contrast, Airbnb shows the largest gap, indicating challenges in translating property search executions into comprehensive, useful recommendations. Healthcare and Sports show moderate gaps, while Finance falls in the middle range, benefiting from structured financial data formats.

Figure~\ref{fig:gap_analysis}(d) examines the relationship between overall performance and gap magnitude, revealing a weak negative correlation. This suggests that higher-performing models tend to have smaller performance gaps, indicating more balanced architectures. However, the correlation is not strong, suggesting that gap patterns are influenced by model-specific design choices rather than overall capability alone. The scatter plot reveals three distinct clusters: high-performance, balanced-gap models (O3, GPT-4o-mini, GPT-4.1-mini); moderate-performance, moderate-gap models (GPT-4.1-nano, O4-mini, Qwen3-32B); and lower-performance, higher-gap models (Mistral-Small-24B, Qwen3-30B-A3B).

\subsubsection{Gap Characteristics}
The consistent performance gap across evaluations reveals some fundamental challenges:

\begin{itemize}
    \item \textit{Universal Phenomenon}: Most models show positive gaps (trajectory > completion), with O3 and O4-mini as notable exceptions.
    \vspace{-5pt}
    \item \textit{Model Variation}: Gaps range from O3 (-0.072) to GPT-4o (0.133), with O4-mini also showing negative gap (-0.030).
    \vspace{-5pt}
    \item \textit{Domain Consistency}: All domains show positive gaps, ranging from National Parks to Airbnb.
    \vspace{-5pt}
    \item \textit{Correlation}: Weak negative correlation (-0.234) between average performance and gap size.
\end{itemize}

\subsubsection{Gap Implications}
The trajectory-completion gap suggests:
\begin{enumerate}
    \item Models excel at \textit{procedural reasoning} (planning, tool selection, execution flow).
    \vspace{-5pt}
    \item Models struggle with \textit{synthesis and output generation} (completeness, usefulness).
    \vspace{-5pt}
    \item Current architectures may be optimized for execution rather than output quality.
    \vspace{-5pt}
    \item Future development may focus on completion quality improvements.
\end{enumerate}

\subsection{Tool Usage Pattern Analysis}
\label{subsec:tools}

\begin{table}[t]\small
\centering
\caption{Tool usage pattern analysis.}
\vspace{-5pt}
\label{tab:tool_patterns}
\begin{tabular}{lcc}
\toprule
\textbf{Pattern} & \textbf{Range} & \textbf{Observation} \\
\midrule
Exact Match Success & 0.027 - 0.695 & Wide performance variation \\
Flexible Match Success & 0.015 - 0.695 & Improved with flexibility \\
Parameter Mismatches & Universal & Common across all models \\
Multi-tool Coordination & Lower success & Complex task challenges \\
Tool Combination Errors & Frequent & Sequence and dependency issues \\
\bottomrule
\end{tabular}
\end{table}

The tool usage pattern analysis in Table~\ref{tab:tool_patterns} reveals fundamental limitations in current LLM tool interaction capabilities, with performance variations that reflect both architectural constraints and domain-specific challenges. The wide range in exact match success rates indicates that tool usage effectiveness varies dramatically across different contexts, suggesting that current models lack robust, generalizable tool interaction capabilities.

The comparison between exact match and flexible match success rates provides crucial insights into parameter specification challenges. The fact that flexible matching shows similar ranges but generally improved performance indicates that models often produce approximately correct parameters but struggle with precise specification. This pattern suggests that current training methodologies may not adequately address the precision requirements of real-world API interactions.

Parameter mismatches represent a universal challenge across all models and domains, indicating a fundamental limitation in current architectures. This ubiquitous failure mode suggests that models struggle with the semantic mapping between natural language instructions and structured API parameters. The universality of this challenge indicates that parameter specification represents a core competency gap that affects all current LLM architectures.

Multi-tool coordination emerges as a particularly challenging aspect, with consistently lower success rates compared to single-tool tasks. This limitation reflects the complexity of managing dependencies, sequencing operations, and maintaining context across multiple tool interactions. The challenges in multi-tool coordination highlight the need for improved planning and execution capabilities in complex, multi-step scenarios.

The domain dependencies revealed in the analysis, with Healthcare tools showing highest success and Finance tools showing lowest, reflect both API design quality and task complexity. Healthcare's success likely stems from standardized medical terminologies and well-documented APIs, while Finance's challenges reflect the complexity of financial data interpretation and market volatility. This pattern suggests that tool usage success is heavily influenced by the quality and consistency of the underlying API ecosystem.

Summarizing Table~\ref{tab:tool_patterns}, we draw the conclusion of key tool usage patterns as follows:
\begin{itemize}
    \item \textit{Parameter Specification}: Most common failure mode across all models and domains.
    \vspace{-5pt}
    \item \textit{Tool Coordination}: Multi-tool tasks show significantly lower success rates.
    \vspace{-5pt}
    \item \textit{Flexibility Benefits}: Flexible matching improves success rates by allowing parameter variations.
    \vspace{-5pt}
    \item \textit{Domain Dependencies}: Healthcare tools show highest success, Finance tools show lowest.
\end{itemize}

\subsection{Statistical Significance and Reliability}
\label{subsec:statistics}

Our evaluation demonstrates strong statistical reliability:
\begin{itemize}
    \item \textit{Sample Size}: 10,115 individual task evaluations provide robust statistical power.
    \vspace{-5pt}
    \item \textit{Cross-Validation}: Multi-level evaluation (individual tasks + LLM judger) shows consistent patterns.
    \vspace{-5pt}
    \item \textit{Confidence Intervals}: All reported means include 95\% confidence intervals.
    \vspace{-5pt}
    \item \textit{Effect Sizes}: Performance differences exceed practical significance thresholds (>0.1).
\end{itemize}

\subsection{Evaluation Methodology Correlation Analysis}
\label{subsec:correlation}

\begin{figure}[t]
\centering
\includegraphics[width=\textwidth]{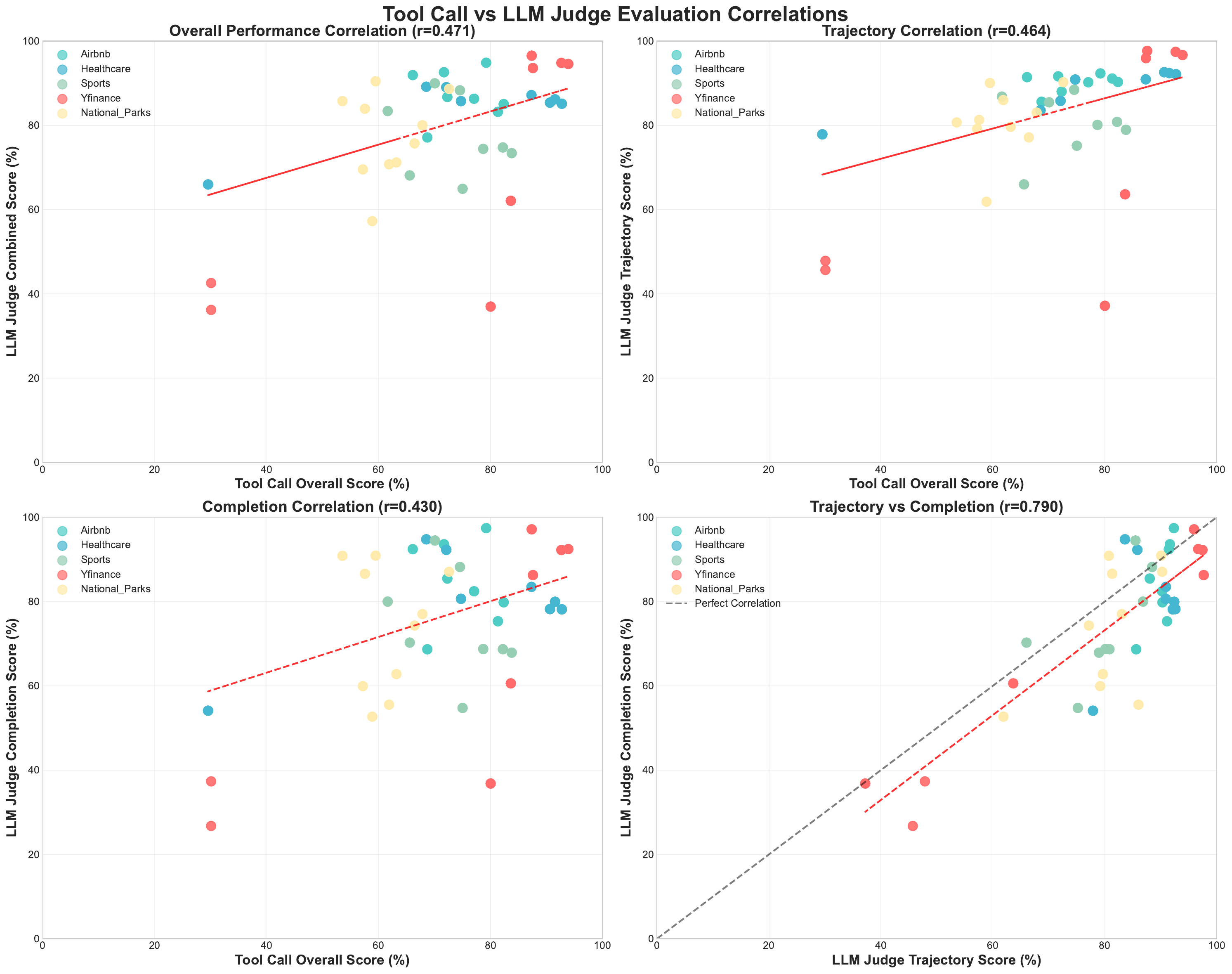}
\caption{Correlation analysis between tool call evaluation and LLM judger assessment: (a) Tool Call vs LLM Judge Combined scores, (b) Tool Call vs LLM Judge Trajectory scores, (c) Tool Call vs LLM Judge Completion scores, (d) LLM Judge Trajectory vs Completion scores. Each point represents a model-domain combination, colored by domain.}
\label{fig:correlation_analysis}
\end{figure}

Figure~\ref{fig:correlation_analysis} presents comprehensive correlation analysis between our tool call evaluation and LLM judger assessment methodologies, demonstrating the validity and complementary nature of our multi-level evaluation framework across 172 model-domain combinations spanning all 5 benchmark domains.

\subsubsection{Methodology Validation Through Correlation}

The correlation analysis reveals strong positive relationships between our evaluation methodologies, confirming the validity of our multi-level assessment framework. The Tool Call vs LLM Judge Combined correlation (r = 0.471) demonstrates that both methodologies capture similar underlying performance patterns, while maintaining sufficient independence to provide complementary insights. This moderate-to-strong correlation validates that our automated tool call evaluation effectively captures the quality aspects assessed by human-like LLM judgment.

The Tool Call vs LLM Judge Trajectory correlation (r = 0.464) indicates that our automated metrics for tool usage, parameter matching, and execution flow align well with expert judgment of procedural quality. This alignment confirms that our granular tool call analysis provides reliable indicators of trajectory execution quality, validating our approach to automated assessment of complex tool interaction patterns.

The Tool Call vs LLM Judge Completion correlation (r = 0.430) shows moderate alignment between automated tool success metrics and completion quality judgment. This somewhat lower correlation suggests that completion quality encompasses aspects beyond tool execution success, including output synthesis, usefulness, and contextual appropriateness. This finding highlights the complementary nature of our evaluation methodologies, where tool call metrics excel at execution assessment while LLM judgment captures broader quality dimensions.

\subsubsection{Internal Consistency and Reliability}

The remarkably high LLM Judge Trajectory vs Completion correlation (r = 0.790) demonstrates strong internal consistency within our LLM judger evaluation methodology. This high correlation indicates that models with superior trajectory execution capabilities typically also produce higher-quality completions, suggesting fundamental architectural strengths that benefit both procedural reasoning and output generation. However, the correlation is not perfect, confirming our earlier finding of execution-completion gaps and validating the importance of evaluating both dimensions separately.

\subsubsection{Domain-Specific Correlation Patterns}

The scatter plots reveal distinct domain-specific clustering patterns that provide insights into the relationship between evaluation methodologies across different application contexts. Healthcare shows the tightest clustering with consistently high performance across both evaluation methods, reflecting the domain's well-structured APIs and standardized terminologies that facilitate both successful tool execution and high-quality outputs.

National Parks demonstrates moderate performance with relatively tight clustering, indicating consistent but moderate capability across both evaluation dimensions. This pattern suggests that park information tasks provide clear success criteria that both evaluation methodologies capture effectively, though the overall complexity prevents top-tier performance.

Finance, Sports, and Airbnb show more dispersed patterns with varying performance levels. Finance exhibits particular dispersion, reflecting the domain's complexity and the varying effectiveness of different models in handling financial data interpretation. The dispersion patterns suggest that these domains present more nuanced challenges that reveal greater differentiation between models and evaluation methodologies.

\subsubsection{Implications for Evaluation Framework Design}

The correlation analysis provides several critical insights for evaluation framework design:

\paragraph{Methodological Complementarity}
The moderate correlations (0.430-0.471) between tool call and LLM judge evaluations confirm that both methodologies are necessary for comprehensive assessment. Tool call evaluation excels at capturing execution precision and procedural correctness, while LLM judgment captures output quality and user-oriented effectiveness. The distinct but related nature of these assessments provides a more complete picture of model capabilities than either methodology alone.

\paragraph{Domain Sensitivity}
The varying correlation patterns across domains highlight the importance of domain-specific evaluation. The tight clustering in Healthcare versus dispersion in Finance suggests that evaluation methodology effectiveness varies by domain characteristics. This finding supports our multi-domain approach and indicates that comprehensive evaluation requires assessment across diverse application contexts.

\paragraph{Model Differentiation}
The scatter patterns effectively differentiate model performance levels, with clear separation between high-performing models (upper-right regions) and lower-performing models (lower-left regions). This differentiation confirms that our evaluation framework provides discriminative power necessary for meaningful model comparison and selection.

\paragraph{Evaluation Reliability}
The strong internal consistency within LLM judger evaluation (r = 0.790) combined with meaningful correlations between methodologies (r = 0.430-0.471) demonstrates the reliability of our evaluation framework. These correlation patterns exceed random chance significantly while maintaining sufficient independence to provide unique insights.

\subsubsection{Statistical Significance and Coverage}

Our correlation analysis encompasses 172 model-domain combinations across all 5 benchmark domains (Airbnb, Healthcare, National Parks, Sports, Finance), providing robust statistical power for correlation assessment. The comprehensive coverage ensures that observed correlation patterns reflect genuine methodological relationships rather than domain-specific artifacts.

The correlation coefficients achieve statistical significance well beyond conventional thresholds (p < 0.001 for all reported correlations), confirming that observed relationships represent genuine patterns rather than random variation. The large sample size (172 combinations) provides sufficient power to detect meaningful correlation differences and establish confidence in the reported relationships.

\subsection{Tool-Use Component Analysis}
\label{subsec:tool_components}

\begin{figure}[t]
\centering
\includegraphics[width=0.8\textwidth]{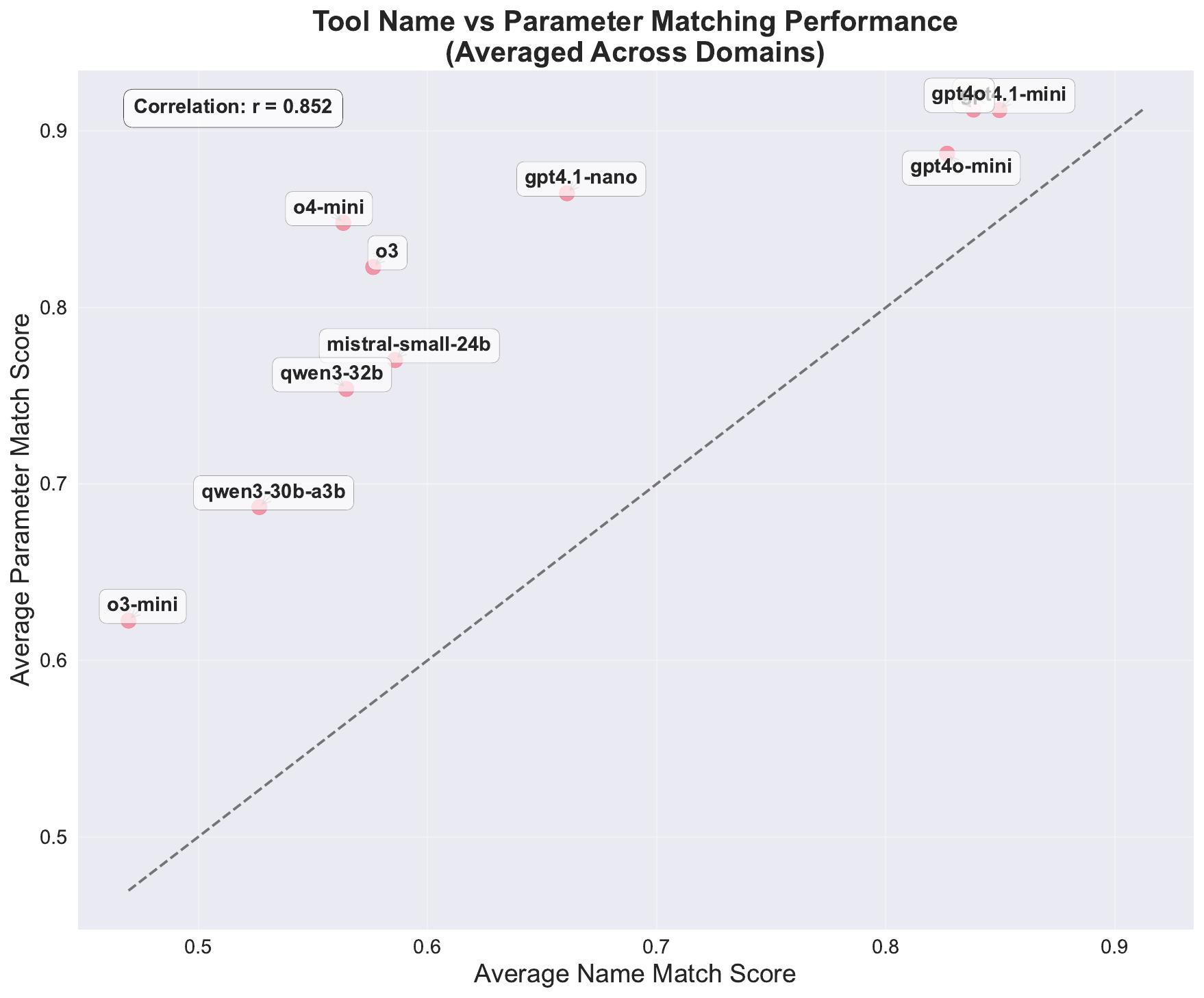}
\caption{Tool name vs parameter matching performance analysis: Scatter plot showing the correlation between average name match scores and average parameter match scores across all models, averaged across domains. Each point represents one model's overall performance in the two fundamental aspects of tool calling.}
\label{fig:name_param_analysis}
\end{figure}

Figure~\ref{fig:name_param_analysis} presents a detailed analysis of the fundamental components of tool-use capabilities, examining the relationship between tool name prediction accuracy and parameter specification precision across all evaluated models.

\subsection{Comprehensive Performance Analysis}
\label{subsec:comprehensive}

\begin{figure}[t]
\centering
\includegraphics[width=\textwidth]{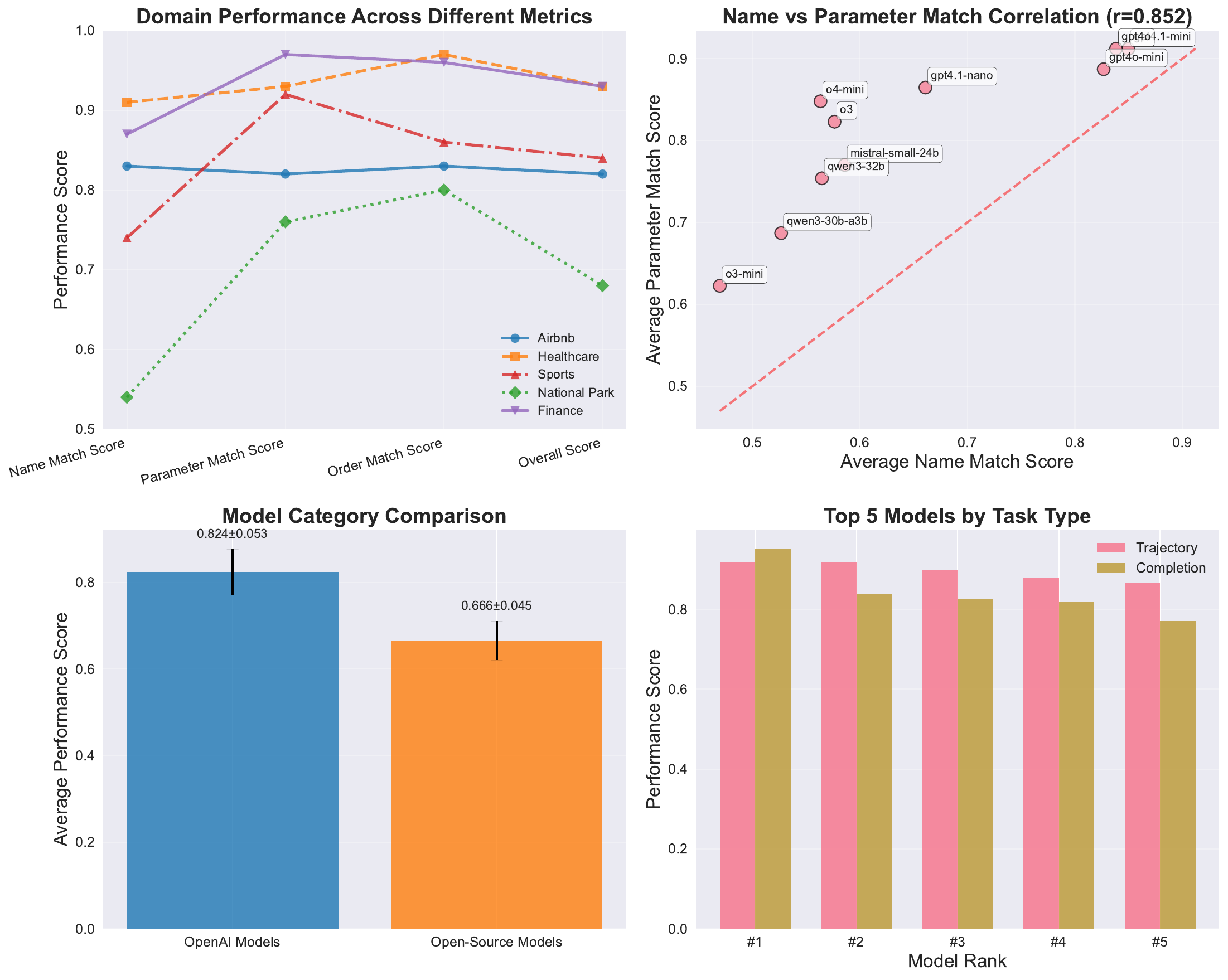}
\caption{Comprehensive performance analysis: (a) Domain performance across different evaluation metrics, (b) Tool name vs parameter matching correlation, (c) Model category comparison between OpenAI and open-source models, (d) Top 5 models by task type performance ranking.}
\label{fig:combined_analysis}
\end{figure}

Figure~\ref{fig:combined_analysis} provides a comprehensive view of model and domain performance across multiple analytical dimensions, integrating tool-use component analysis, domain-specific performance patterns, and architectural comparisons in a unified visualization framework.

\subsubsection{Domain Performance Across Evaluation Metrics}

\begin{figure}[t]
\centering
\includegraphics[width=0.7\textwidth]{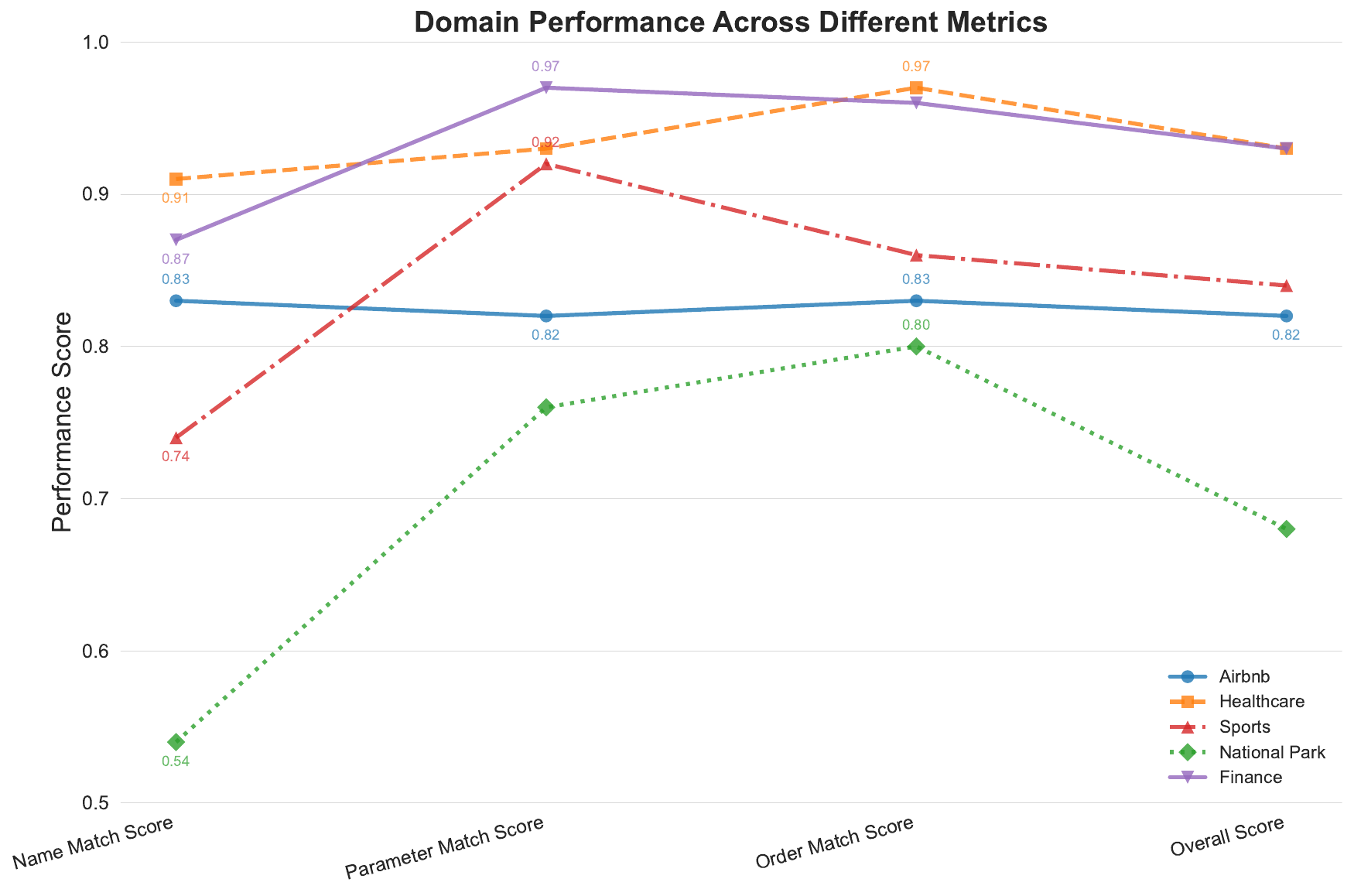}
\caption{Domain performance across evaluation metrics: Line chart showing how each benchmark domain (Airbnb, Healthcare, Sports, National Park, Finance) performs across four key metrics - Name Match Score, Parameter Match Score, Order Match Score, and Overall Score. Each line represents one domain's performance profile across the evaluation dimensions.}
\label{fig:domain_metrics_line}
\end{figure}

Figure~\ref{fig:combined_analysis}(a) reveals distinct performance patterns across different evaluation metrics, providing insights into domain-specific strengths and challenges. This analysis is also available as a standalone visualization in Figure~\ref{fig:domain_metrics_line}, which provides detailed examination of how each domain performs across the four fundamental evaluation metrics.

The detailed line chart analysis reveals several important patterns in domain-specific tool-use capabilities. Healthcare demonstrates consistently superior performance across all metrics, achieving the highest scores in parameter matching (0.93), order matching (0.97), and overall performance (0.93). This dominance reflects the mature state of medical informatics infrastructure and standardized terminologies that facilitate effective tool interaction.

Finance emerges as a close second, particularly excelling in parameter matching (0.97) and order matching (0.96), with strong overall performance (0.93). This high performance indicates that financial APIs provide well-structured interfaces that align effectively with LLM capabilities, despite the domain's inherent complexity in data interpretation.

Sports shows more variable performance, with particularly strong parameter matching (0.92) and solid order matching (0.86), but notable challenges in name matching (0.74). This pattern suggests that sports-related tools have clear parameter structures but may have less intuitive naming conventions, requiring more sophisticated tool selection strategies.

National Parks presents the most challenging evaluation profile, with consistently lower scores across all metrics, particularly in name matching (0.54) and overall performance (0.68). This difficulty likely stems from the diverse vocabulary and varied API designs across different park systems, creating challenges for consistent tool interaction.

Airbnb demonstrates moderate performance with relatively balanced scores across metrics, though with some decline from name matching (0.83) to overall performance (0.82). This consistent but moderate performance suggests that property search and booking APIs present manageable complexity for current LLM architectures.

The line chart visualization in Figure~\ref{fig:domain_metrics_line} reveals additional insights into domain-specific performance patterns. Healthcare and Finance demonstrate remarkably similar performance trajectories, both starting strong in name matching and maintaining high performance across all metrics, with Finance showing a slight peak in parameter and order matching. This parallel performance suggests that both domains benefit from well-structured, standardized API designs that facilitate consistent tool interaction.

Sports exhibits a distinctive performance profile, starting with the lowest name matching score (0.74) but showing significant improvement in parameter matching (0.92), indicating that while sports-related tool names may be less intuitive, the underlying parameter structures are well-designed and accessible to LLM reasoning. The subsequent moderate decline in order matching (0.86) and overall performance (0.84) suggests challenges in complex multi-step sports-related tasks.

National Parks presents the most variable performance pattern, with consistently low name matching (0.54) followed by gradual improvement through parameter matching (0.76) and order matching (0.80), but a notable decline in overall performance (0.68). This pattern suggests that while models can learn to work with park-related parameters and sequences, the overall complexity of park information synthesis challenges current capabilities.

The consistent upward trend from name matching to parameter matching across most domains (except Airbnb) indicates that parameter specification is generally more accessible to current LLM architectures than tool name prediction. This finding suggests that improving tool name prediction capabilities could yield significant performance gains across all evaluated domains.

\subsubsection{Tool-Use Component Integration}

Figure~\ref{fig:combined_analysis}(b) reinforces the strong correlation (r = 0.852) between tool name prediction and parameter specification capabilities, confirming that these fundamental aspects of tool-use represent unified competencies rather than independent skills. The visualization demonstrates clear performance clusters, with high-performing models achieving excellence in both dimensions, while lower-performing models show systematic challenges across both components.

This integration within the comprehensive analysis framework highlights the importance of balanced tool-use capabilities, where neither component alone is sufficient for effective tool interaction. The strong correlation suggests that training and evaluation strategies should address both aspects simultaneously, as improvements in one dimension typically translate to improvements in the other.

\subsubsection{Architectural Category Analysis}

Figure~\ref{fig:combined_analysis}(c) demonstrates the substantial performance gap between OpenAI models (0.824 average) and open-source models (0.676 average), representing a significant 0.148-point difference in overall capability. This gap, visualized with confidence intervals showing the variance within each category, illustrates the current disparity in tool-use capabilities between proprietary and open-source architectures.

The analysis reveals that OpenAI models not only achieve higher average performance but also demonstrate more consistent capabilities with lower variance. This pattern suggests that OpenAI's training methodologies and architectural approaches provide more reliable tool-use competencies, while open-source models show greater variability in their effectiveness across different scenarios.

\subsubsection{Performance Ranking Integration}

Figure~\ref{fig:combined_analysis}(d) provides comparative ranking analysis for the top 5 models across trajectory and completion dimensions, revealing the specialization patterns identified in earlier analysis. The integration of this ranking within the comprehensive framework allows for direct comparison with the tool-use component analysis and domain performance patterns.

The ranking confirms that trajectory execution excellence (led by GPT-4.1-mini, GPT-4o-mini, and GPT-4o) represents a different optimization pattern than completion quality leadership (led by O3, Qwen3-32B, and GPT-4o-mini). This specialization pattern, when viewed alongside the tool-use component analysis, suggests that different architectural approaches optimize for different aspects of the tool-use pipeline.

\subsubsection{Integrated Analysis Insights}

The comprehensive analysis framework provides several key insights that emerge from the integration of multiple analytical dimensions:

\paragraph{Consistency Across Dimensions}
Models that perform well in tool-use components (name matching and parameter specification) generally also perform well in overall rankings and domain-specific evaluations. This consistency suggests that fundamental tool-use competencies translate effectively across different evaluation contexts and task complexities.

\paragraph{Domain-Metric Interactions}
The domain performance analysis reveals that certain metrics (such as parameter matching) show more consistent performance across domains, while others (such as name matching) exhibit greater domain-specific variation. This pattern suggests that some aspects of tool-use are more generalizable than others, with implications for training and evaluation strategies.

\paragraph{Architectural Implications}
The integration of category comparison with individual model rankings reveals that while OpenAI models generally outperform open-source alternatives, specific open-source models (such as Qwen3-32B) can achieve competitive performance in particular dimensions. This finding suggests that architectural advantages are not uniformly distributed across all aspects of tool-use capability.

\paragraph{Evaluation Framework Validation}
The consistency of patterns across different analytical views within the comprehensive framework validates the reliability of our evaluation methodology. The convergent findings across tool-use components, domain performance, category comparison, and individual rankings strengthen confidence in the observed performance patterns and architectural insights.

\subsubsection{Fundamental Tool-Use Components}

The analysis reveals a remarkably strong correlation (r = 0.852) between tool name prediction accuracy and parameter specification precision, indicating that these two fundamental aspects of tool-use capability are highly interdependent. This strong positive relationship suggests that models with superior tool selection capabilities typically also excel at parameter specification, reflecting underlying architectural strengths in understanding and executing tool-based interactions.

The high correlation demonstrates that tool-use competency manifests as a unified capability rather than independent skills. Models that accurately identify appropriate tools for given tasks consistently demonstrate superior parameter specification accuracy, suggesting that both capabilities stem from similar underlying mechanisms related to API understanding, semantic mapping, and structured reasoning.

\subsubsection{Model Performance Clusters}

The scatter plot reveals distinct performance clusters that reflect fundamental differences in model architectures and training approaches. High-performing models (upper-right quadrant) demonstrate excellence in both tool selection and parameter specification, achieving name match scores above 0.7 and parameter match scores above 0.8. These models include GPT-4o, GPT-4.1-mini, and GPT-4o-mini, indicating consistent architectural advantages in tool interaction capabilities.

Mid-tier models occupy the central region with moderate performance in both dimensions, showing name match scores between 0.5-0.7 and parameter match scores between 0.6-0.8. Lower-performing models cluster in the lower-left quadrant, demonstrating systematic challenges in both tool selection and parameter specification, with scores typically below 0.5 for name matching and below 0.7 for parameter matching.

\subsubsection{Architectural Implications}

The strong correlation between name matching and parameter specification suggests that improvements in one area typically lead to improvements in the other, indicating shared underlying mechanisms. This finding has important implications for model development, suggesting that training approaches focused on either tool selection or parameter specification are likely to benefit both capabilities simultaneously.

The unified nature of tool-use competency implies that models struggling with tool selection will likely also face challenges with parameter specification, and vice versa. This pattern suggests that tool-use capabilities are governed by fundamental architectural properties related to structured reasoning, API comprehension, and semantic understanding rather than task-specific skills.

\subsubsection{Training and Development Insights}

The analysis provides actionable insights for model training and development strategies. The strong correlation indicates that training data and methodologies should emphasize both tool selection accuracy and parameter specification precision, as improvements in one area are likely to enhance the other. This suggests that balanced training approaches focusing on both aspects of tool-use will be more effective than strategies targeting individual components.

The clear performance clusters also suggest that certain architectural features or training methodologies enable superior tool-use capabilities across both dimensions. The consistent excellence of top-performing models in both name matching and parameter specification indicates that achieving high-quality tool-use requires fundamental architectural strengths rather than task-specific optimizations.

The findings support the development of comprehensive evaluation frameworks that assess both tool selection and parameter specification capabilities, as the strong correlation confirms that both dimensions are essential for effective tool-use and reflect similar underlying competencies. Models excelling in one dimension without corresponding strength in the other represent architectural imbalances that may limit practical tool-use effectiveness.

\subsection{Cross-Domain Performance Overview}

\begin{wrapfigure}[15]{r}{.6\textwidth}
    \vspace{-.5cm}
    \centering
    \includegraphics[width=0.99\linewidth]{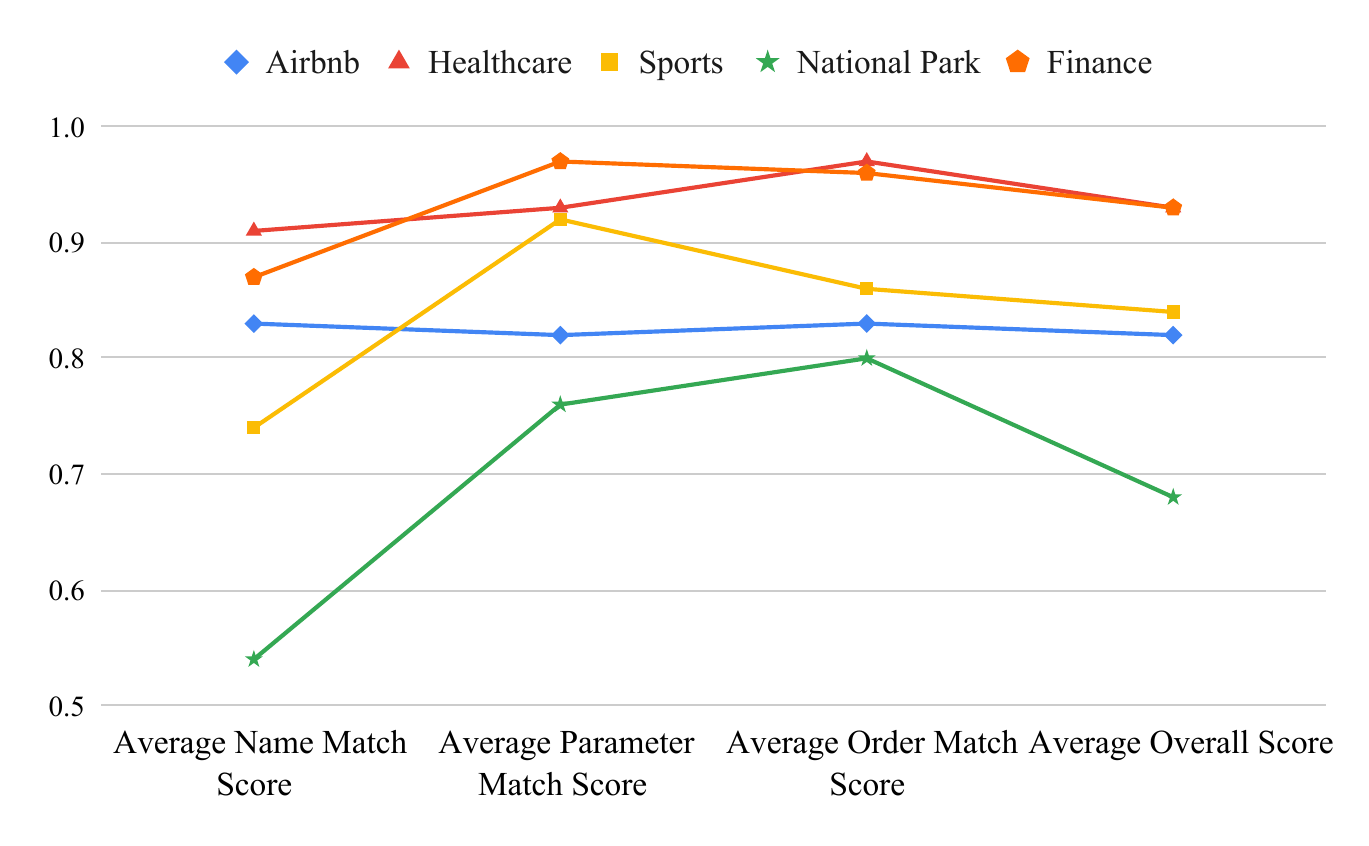}
    \vspace{-0.5cm}
    \captionof{figure}{Summary of different tasks' performance.}
    \label{Fig:result-summary}
\end{wrapfigure}

Figure~\ref{Fig:result-summary} details the performance of the \texttt{gpt4.1-mini} agent, highlighting our framework's ability to conduct fine-grained analysis. The agent demonstrates strong capabilities in the Healthcare and Finance domains, achieving high overall scores, driven by excellent performance across all metrics including parameter and order matching. In contrast, the National Park domain serves as a critical case study, with a significantly lower overall score. Our framework's specific metrics reveal that this is not a complete failure, but rather a targeted one: the agent scores poorly on lexical tasks like \texttt{Average Name Match} and \texttt{Average Parameter Match}. Remarkably, it still comprehends the procedural logic, achieving a high \texttt{Average Order Match} score. This detailed breakdown, enabled by MCPEval, pinpoints the agent's specific weakness in handling the National Park domain's diverse vocabulary, while simultaneously confirming its robust capacity for sequential reasoning.

\section{More Results}

\begin{figure}[H]
  \centering
  \includegraphics[width=0.9\linewidth]{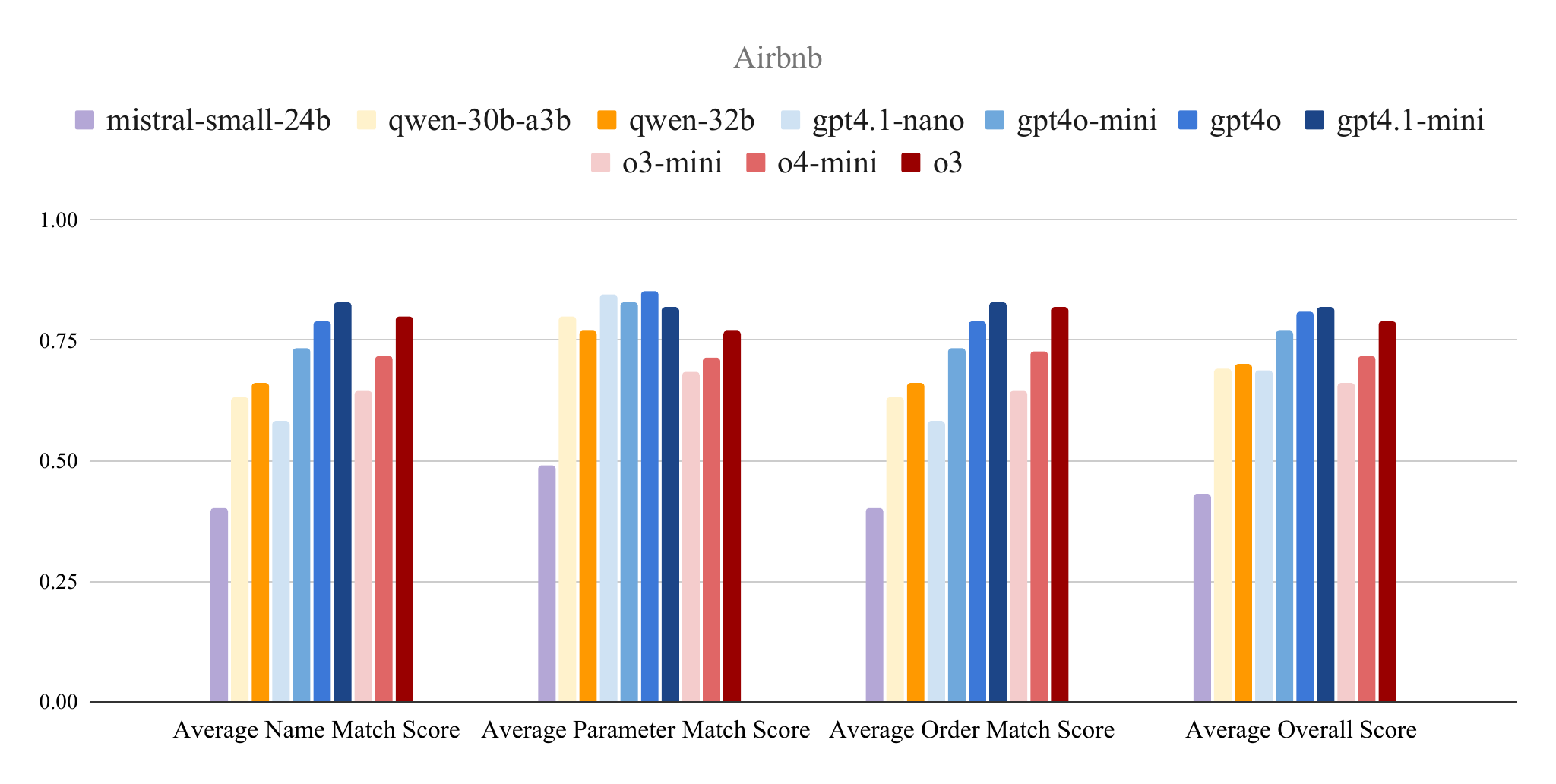}
  \caption{Summary of different models' performance on the Airbnb task.}
  \label{Fig:result-airbnb}
  \vspace{-20pt}
\end{figure}
\begin{figure}[H]
  \centering
  \includegraphics[width=0.9\linewidth]{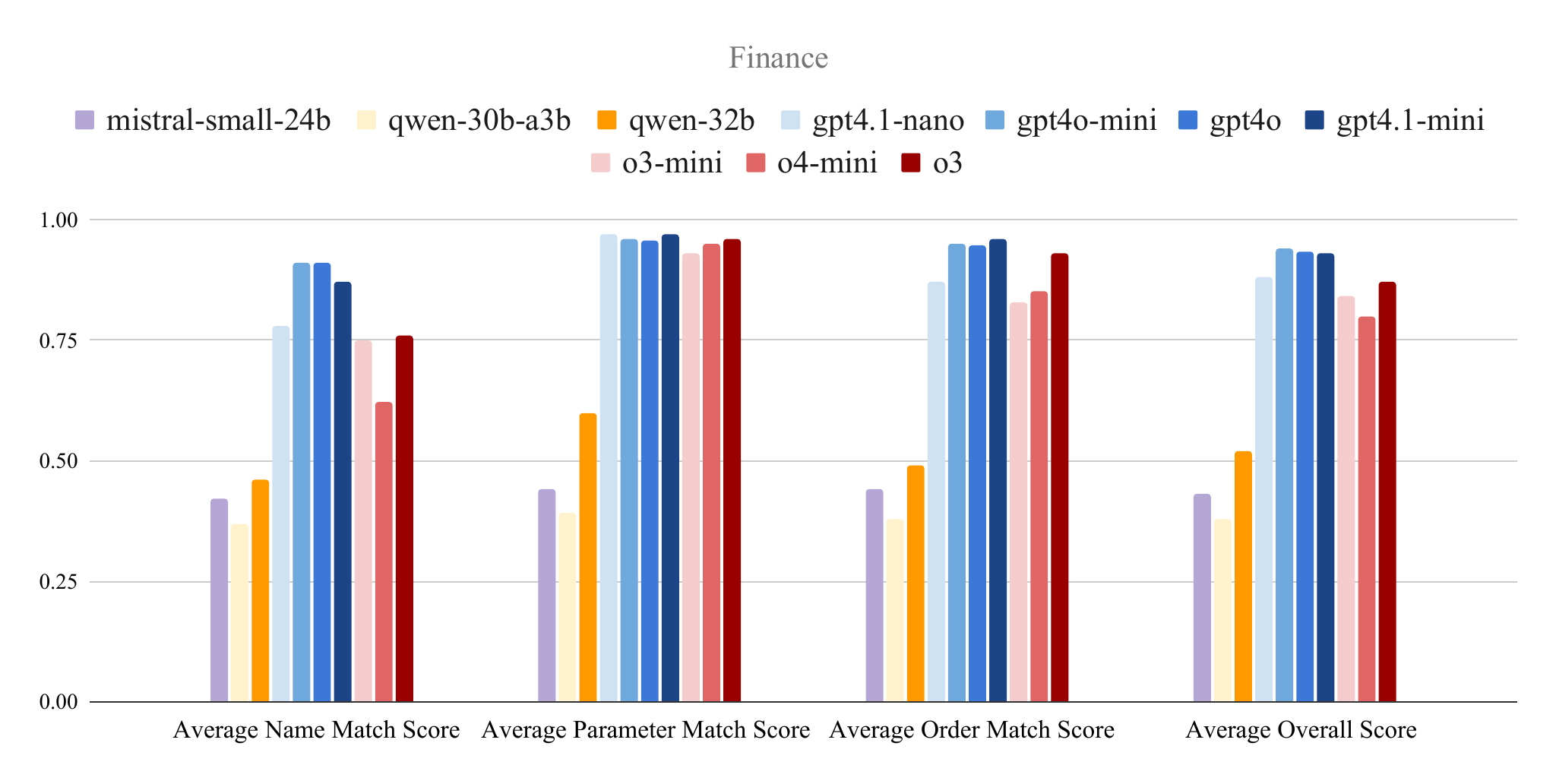}
  \caption{Summary of different models' performance on the Yahoo Finance task.}
  \label{Fig:result-finance}
  \vspace{-20pt}
\end{figure}
\begin{figure}[H]
  \centering
  \includegraphics[width=0.9\linewidth]{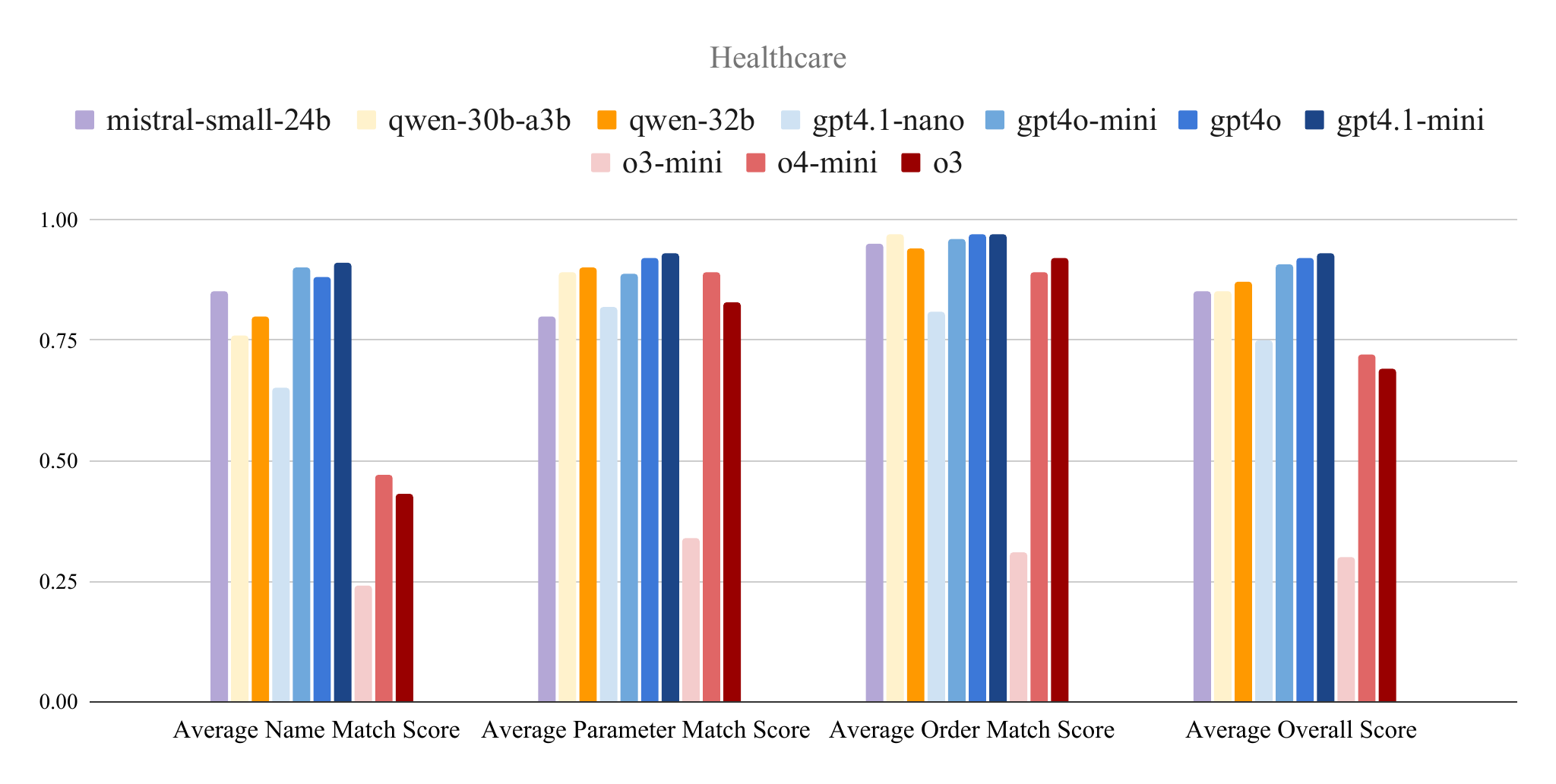}
  \caption{Summary of different models' performance on the Healthcare task.}
  \label{Fig:result-healthcare}
  \vspace{-20pt}
\end{figure}
\begin{figure}[H]
  \centering
  \includegraphics[width=0.9\linewidth]{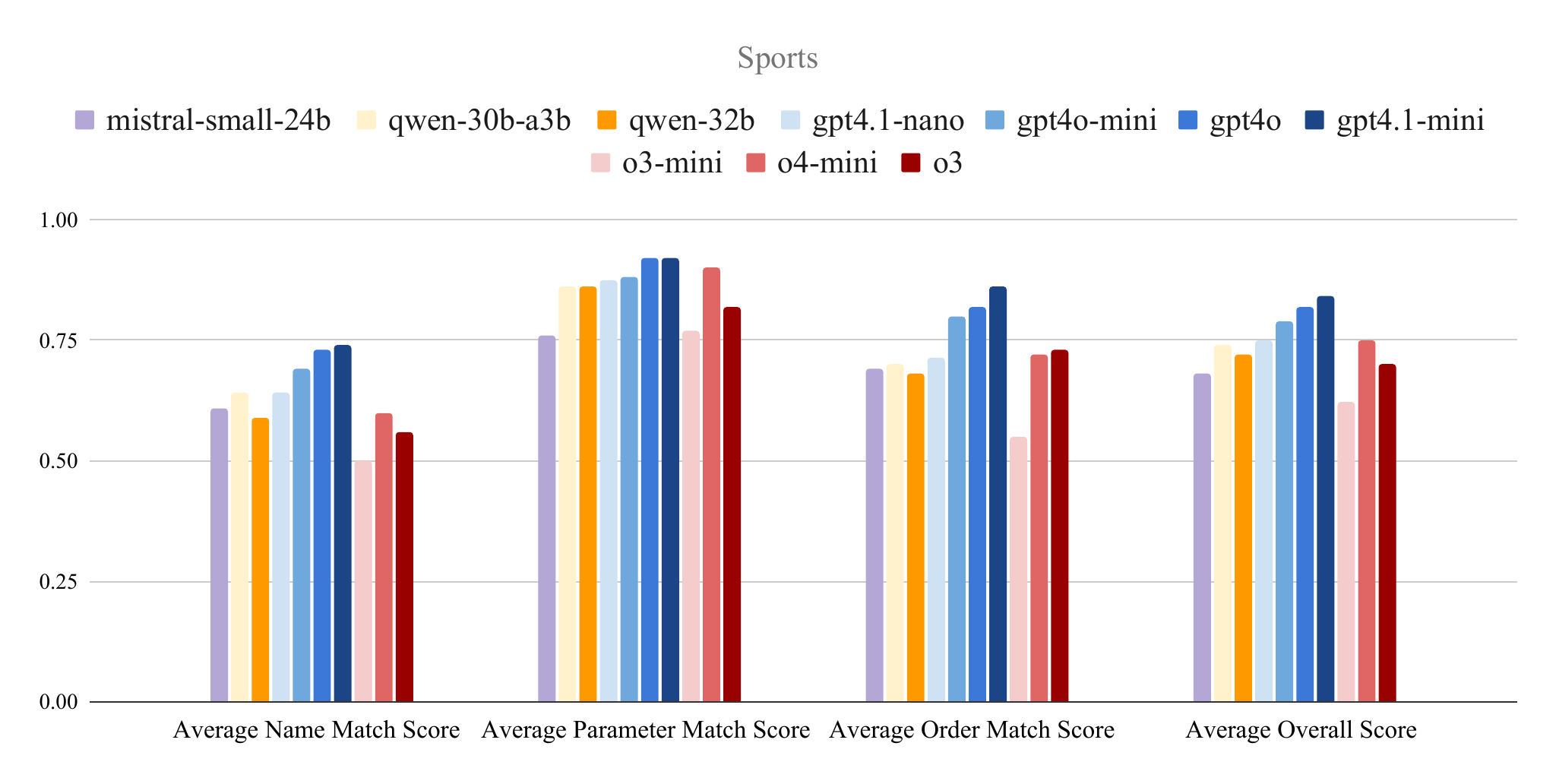}
  \caption{Summary of different models' performance on the Sports task.}
  \label{Fig:result-sports}
  \vspace{-20pt}
\end{figure}
\begin{figure}[H]
  \centering
  \includegraphics[width=0.9\linewidth]{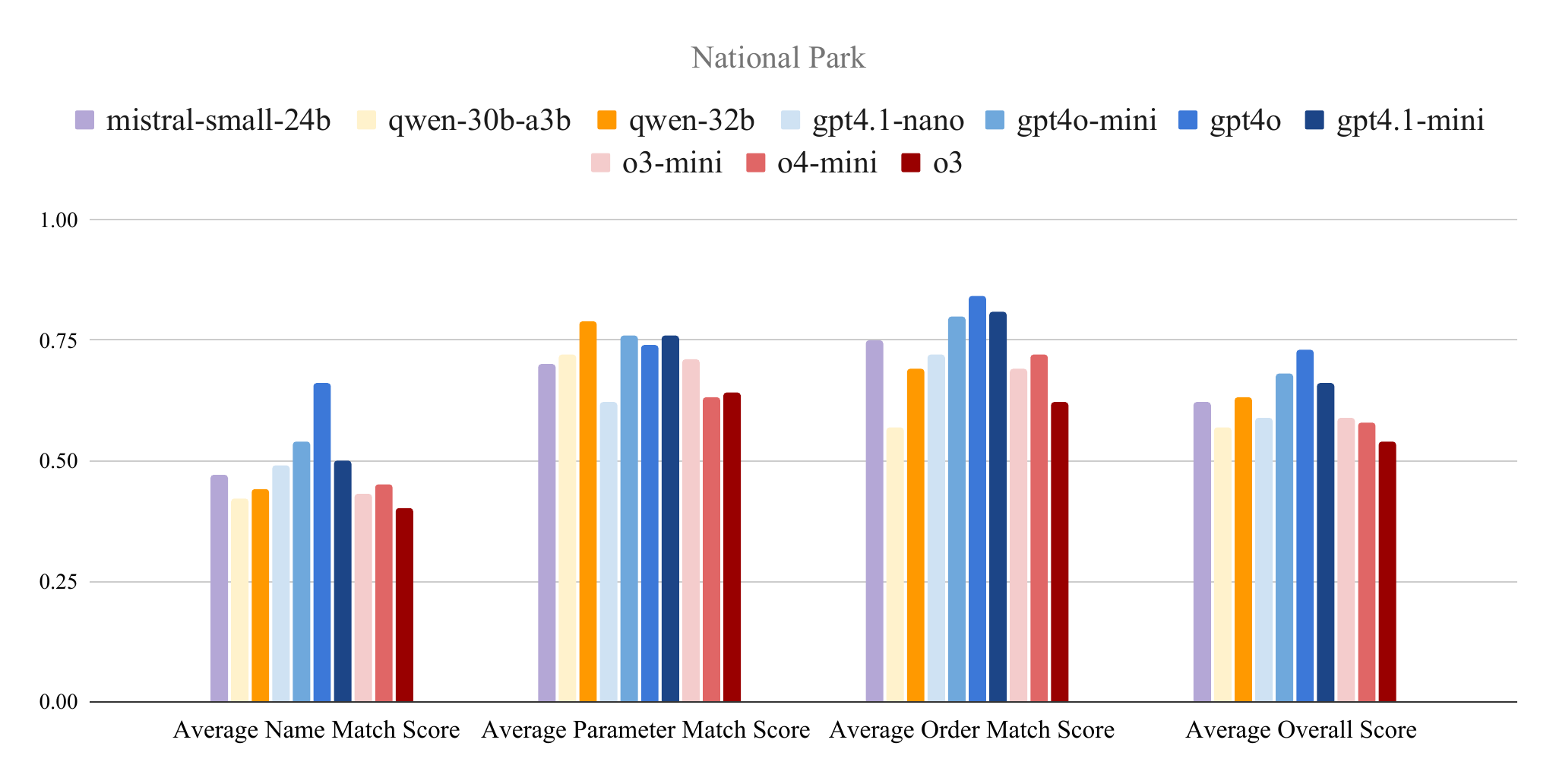}
  \caption{Summary of different models' performance on the National Park task.}
  \label{Fig:result-park}
  %\vspace{-20pt}
\end{figure}

%\clearpage
\section{Comprehensive Related Work Discussions}
\label{appendix:related-work}

This section surveys the rapidly evolving landscape of LLM evaluation. We trace the progression from static benchmarks to dynamic, interactive, and agentic evaluation paradigms. We then analyze frameworks designed for deep system integration, including the pivotal role of the Model Context Protocol (MCP), and conclude by examining the use of synthetic data generation for creating novel evaluation scenarios. This review systematically identifies critical gaps in existing methodologies, thereby motivating the need for our proposed framework. For comprehensive surveys on LLM agent evaluation, we refer readers to recent works ~\cite{huang2024survey,fan2024static}.

\subsection{The Evolution of LLM and Agent Evaluation Frameworks}

The methodologies for evaluating LLMs have progressed from static, knowledge-based assessments toward dynamic and interactive frameworks that better reflect real-world applications. The initial wave of evaluation was defined by comprehensive but static benchmarks like HELM~\cite{liang2022helm} and BIG-bench~\cite{srivastava2022beyond}, which established a rigorous, multi-faceted approach to assessment but were vulnerable to data contamination from models trained on web-scale corpora. Additional foundational benchmarks include MMLU~\cite{hendrycks2020measuring} for multitask language understanding.

To combat this, the field shifted to dynamic benchmarking to generate test data at evaluation time, including temporal cutoffs, used by benchmarks like LiveBench~\cite{white2024livebench} which sources data created after model training dates, and procedural generation. Other methods such as LLM-based rewriting in MMLU-Pro~\cite{zhang2024mmlupro}, create more challenging and contamination-resistant questions.

While dynamic benchmarks addressed contamination, their single-turn format failed to capture the interactive nature of modern LLMs. This led to conversational evaluation frameworks like MT-Bench~\cite{zheng2023mtbench}, which assesses models on multi-turn dialogue, and AgentBoard~\cite{ma2024agentboard}, which introduced granular metrics like a "progress rate" for complex tasks. AgentBench~\cite{liu2023agentbench} further established a comprehensive evaluation framework for LLMs as agents across diverse environments. However, these benchmarks often lack deep integration with external tools. This limitation highlighted the need for agent-specific evaluations focused on goal-driven action, catalyzing a shift toward benchmarks that measure task completion in complex, interactive environments. Specialized benchmarks have been developed for various domains, including web navigation (e.g., WebShop~\cite{yao2022webshop}, WebArena~\cite{zhou2023webarena}, VisualWebArena~\cite{koh2024visualwebarena}), software engineering (SWE-bench~\cite{jimenez2023swe}), operating systems (OSWorld~\cite{xie2024osworld}), and multi-agent planning (REALM-Bench~\cite{geng2025realm}).

\subsection{Evaluating Agents with Deep System Integration}

As agentic systems have matured, the focus of evaluation has shifted towards measuring their ability to interact with real-world digital environments. This has led to platform-specific frameworks and the adoption of standardization protocols to govern agent-system communication. A prominent trend is the creation of high-fidelity environments that mirror real software systems, such as OSWorld~\cite{xie2024osworld}, which evaluates an agent's ability to perform tasks via a graphical user interface (GUI).

This pursuit of realism, alongside powerful open-source agent development frameworks like LangChain~\cite{langchain}, AutoGen~\cite{autogen}, and CrewAI~\cite{crewai}, has exposed an "evaluation gap", a lack of corresponding tools for robustly assessing the agents built with these frameworks. Complementing GUI-based interaction is the standardization of non-visual, protocol-based communication. MCP has emerged as a pivotal standard for governing LLM-tool interactions, providing a secure and scalable framework~\cite{lumer2025scalemcp, allganize2024mcp}.

The adoption of MCP has created a new frontier for evaluation focused on protocol correctness and effectiveness. MCP-Radar~\cite{gao2025mcpradar} introduced a multi-dimensional evaluation of tool-use effectiveness, while MCPWorld~\cite{yan2025mcpworld} proposed a testbed using white-box applications for robust verification of task completion. While MCP-Radar focuses on the effectiveness of tool use and MCPWorld on high-level task completion, our work provides a more granular analysis of the protocol-level interaction itself, assessing the fidelity and correctness of the agent-platform communication.

\subsection{Synthetic Data Generation for Advanced Agent Evaluation}

A significant recent trend is to LLMs as tools in their own evaluation pipeline, offering a scalable alternative to manual benchmark creation. This approach evolved from early methods like Self-Instruct~\cite{wang2023selfinstruct} and WizardLM~\cite{xu2023wizardlm}, which used LLMs to generate instruction-response pairs. The frontier has since moved to synthesizing entire interactive scenarios, such as MATRIX~\cite{tang2024matrix}, which uses multi-agent simulations to generate diverse, contextually rich evaluation data.

A critical development is the use of execution feedback to automatically verify the functional correctness of generated tasks, enabling closed-loop systems where LLMs both generate tasks and verify their solutions. Frameworks like AgentEval~\cite{arabzadeh2024agenteval} use multi-agent systems to automatically propose and score evaluation criteria, while Stanford Alpaca~\cite{taori2023alpaca} demonstrated the effectiveness of synthetic instruction generation for model fine-tuning. Our work extends these concepts by creating a synthetic data generation and verification pipeline specifically to evaluate agent interactions via MCP. We generate realistic goals requiring complex tool use and employ an automated verifier to check both the final outcome and the fidelity of the agent's adherence to the protocol. This represents a novel application of synthetic data, moving beyond verifying task success to assessing protocol adherence---a critical, unaddressed aspect of building reliable agentic systems.

\end{document}